\documentclass{article} % For LaTeX2e
\usepackage{iclr2027_conference,times}

\usepackage{amsmath,amsfonts,bm}

\def\eqref#1{equation~\ref{#1}}
\def\1{\bm{1}}

\DeclareMathAlphabet{\mathsfit}{\encodingdefault}{\sfdefault}{m}{sl}
\SetMathAlphabet{\mathsfit}{bold}{\encodingdefault}{\sfdefault}{bx}{n}

\usepackage{hyperref}
\usepackage{url}
\usepackage{graphicx}
\usepackage{multirow}
\usepackage{caption}
\usepackage{wrapfig}
\usepackage[table]{xcolor}  
\usepackage{amsmath, amssymb, amsfonts}
\usepackage{tabularx}
\usepackage{arydshln}
\usepackage{booktabs}
\usepackage{makecell}
\usepackage{float}
\usepackage{pifont}   
\usepackage[ruled,vlined,linesnumbered]{algorithm2e}  
\newcommand{\cmark}{\ding{51}}
\newcommand{\xmark}{\ding{55}}
\definecolor{darkgreen}{RGB}{0,100,0}
\definecolor{darkred}{RGB}{180,0,0}
\definecolor{darkorange}{RGB}{210,105,30}
\newcommand{\tablestyle}[2]{\setlength{\tabcolsep}{#1}\renewcommand{\arraystretch}{#2}\centering\footnotesize}

\title{LoopQuant: Accurate and Efficient Post-Training Quantization for Looped Transformers}

\author{%
  \makebox[\dimexpr\textwidth-2\tabcolsep\relax][c]{%
    \begin{tabular}{@{}c@{}}
      \textbf{Zhixiong Zhao}\textsuperscript{1,2}\enspace
      \textbf{Zukang Xu}\textsuperscript{2}\enspace
      \textbf{Xing Hu}\textsuperscript{2}\enspace
      \textbf{Lifeng Liu}\textsuperscript{1}\enspace
      \textbf{Guangyu Sun}\textsuperscript{1}\enspace
      \textbf{Dawei Yang}\textsuperscript{2}\\[0.7em]
      {\normalfont\textsuperscript{1}School of Integrated Circuits, Peking University}\\
      {\normalfont\textsuperscript{2}Houmo AI}
    \end{tabular}%
  }%
}

\iclrfinalcopy 
\begin{document}

\maketitle

\begin{abstract}
Looped Language Models (LoopLMs) improve parameter efficiency by recursively reusing shared Transformer modules, yet their repeated computation over evolving recurrent states poses new challenges for low-bit deployment.
Existing post-training quantization (PTQ) methods, largely designed for feed-forward Transformers, fail to account for cross-loop activation dynamics, heterogeneous loop sensitivity, and recursive error propagation, leading to substantial performance degradation at low precision.
To address these challenges, we propose LoopQuant, a loop-aware PTQ framework for accurate and efficient low-bit quantization of LoopLMs.
LoopQuant is built on three core components: 1) Cross-Loop Distribution Canonicalization (CLDC) learns a shared quantization-friendly representation across recurrent states while preserving parameter sharing; 2) Interaction-Aware Loop-wise Precision Allocation (ILPA) allocates activation precision under a fixed bit budget by modeling task-level loop sensitivity and cross-loop interactions; and 3) Loop-Conditioned Task Calibration (LCTC) further aligns quantized inference with the low-bit recursive trajectory through lightweight task-space calibration.
Extensive experiments across LoopLM architectures demonstrate that LoopQuant consistently outperforms existing low-bit PTQ methods.
On Ouro-2.6B, LoopQuant under W4A4 surpasses all W4A8 baselines in performance, while reducing model memory by 65.3\% with only 3.4\% additional storage overhead over naive INT4 and achieving a 1.7$\times$ decoding speedup over BF16.
These results demonstrate a strong balance among accuracy, compression, and inference efficiency, providing a practical and deployment-friendly solution for low-bit LoopLM inference.
The code is available at https://github.com/Kishon-zzx/LoopQuant.
\end{abstract}

\section{Introduction}
\label{sec:intro}

Transformer-based large language models (LLMs) have demonstrated strong capabilities in language understanding, code generation, and complex reasoning. However, these gains come with larger model sizes, higher inference costs, and increased memory access, making deployment challenging in memory- and bandwidth-constrained environments. Rather than simply scaling model size, Looped Language Models (LoopLMs) recursively reuse the same Transformer modules across multiple loops~\citep{zhu2026scalinglatentreasoninglooped}, increasing effective computation depth under a nearly fixed parameter budget. This parameter sharing reduces storage while maintaining strong representation and reasoning capabilities, making LoopLMs attractive for constrained deployment. However, shared modules are repeatedly executed, with weights and intermediate activations frequently accessed across loops. Low-bit quantization of both weights and activations is therefore important for reducing computation and data movement costs. Achieving efficient low-bit quantization while preserving recursive reasoning capabilities remains a key challenge for practical LoopLM deployment.

\begin{figure*}[t!]
    \centering
    \includegraphics[width=0.95\textwidth]{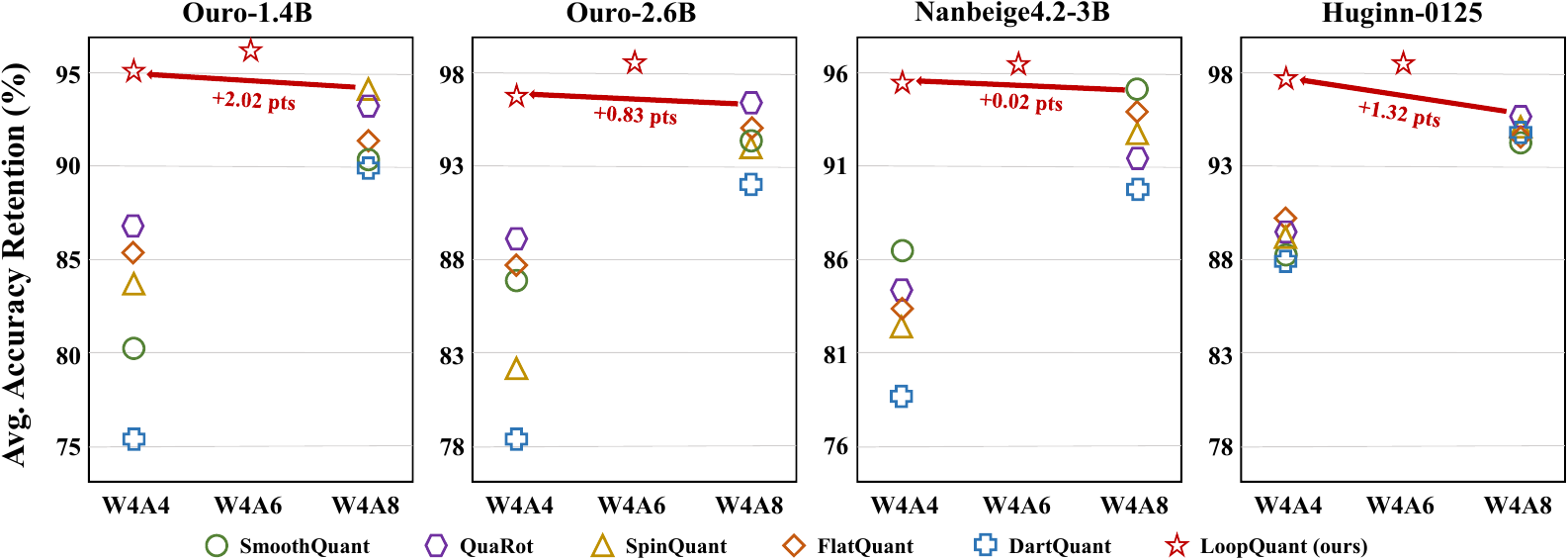}
    \caption{Accuracy retention of PTQ methods on four LoopLMs under different W/A precisions. LoopQuant consistently preserves higher accuracy, with W4A4 surpassing baselines at W4A8.}
    \label{fig:accuracy_retention}
    %\vspace{-0.4cm}
\end{figure*}

Post-training quantization (PTQ) enables efficient LLM deployment by reducing storage and computation costs without expensive retraining. Existing methods employ weight reconstruction, activation clipping, channel scaling, rotation transformations~\citep{xiao2023smoothquant,ashkboos2024quarot}, and mixed-precision allocation~\citep{saxena2024resq} to enable low-bit weight and activation quantization for conventional Transformers. However, most target feed-forward Transformers, where layers are executed sequentially and quantization is calibrated from relatively stable layer-wise activation statistics, typically with local reconstruction objectives. LoopLMs break this assumption. The same Transformer modules are repeatedly invoked across loops, while evolving recurrent states expose them to different activation distributions. Quantization perturbations from earlier loops also propagate through recursive computation, altering subsequent hidden states and quantization behavior. LoopLM quantization must therefore account for cross-loop activation dynamics, loop-dependent quantization sensitivity, and recursive error propagation. As shown in Fig.~\ref{fig:accuracy_retention}, directly applying existing PTQ methods can cause substantial performance degradation at low bit-widths.

To further understand the core challenges of low-bit PTQ for LoopLMs, we identify three key issues.
\ding{182} \textbf{Activation distributions vary substantially across loops.} As hidden states evolve, the same shared module receives different activation distributions across loops, with varying dynamic ranges, outlier magnitudes, and outlier channels. Static quantization parameters calibrated on a single distribution may fail to accommodate all recurrent states, while separate configurations for each loop introduce additional storage and deployment overhead.
\ding{183} \textbf{Loops exhibit heterogeneous importance and quantization sensitivity.} Although all loops share the same Transformer parameters, they operate at different recurrent stages and contribute differently to the final prediction. Quantization errors introduced in earlier loops can propagate through later computation. Uniform activation precision may therefore waste the bit budget, while allocation based only on local quantization errors cannot reflect the true impact of each loop on the final output.
\ding{184} \textbf{Local quantization objectives are misaligned with recursive task behavior.} Even when the local quantization error is small at one loop, low-bit computation can shift the hidden state and alter the input distributions of subsequent loops. Such deviations can accumulate along the recursive trajectory and cause substantial output errors. Accurate LoopLM quantization therefore requires joint modeling of cross-loop activation dynamics and loop sensitivity, followed by calibration along the actual low-bit recursive trajectory.

To this end, we propose \textbf{LoopQuant}, a loop-aware PTQ framework for accurate low-bit LoopLM inference.
\ding{182} \textbf{Cross-Loop Distribution Canonicalization (CLDC)} learns a shared quantization-friendly representation across recurrent states through orthogonal-scaling transformation, robust cross-loop synchronization, and group-aware channel reordering, while preserving parameter sharing.
\ding{183} \textbf{Interaction-Aware Loop-wise Precision Allocation (ILPA)} allocates A4/A8 activation precision under a fixed bit budget by estimating task-level loop sensitivity and modeling cross-loop interactions induced by recursive error propagation.
\ding{184} \textbf{Loop-Conditioned Task Calibration (LCTC)} further calibrates quantization behavior along the actual low-bit recursive trajectory through task-space defect transport and Fisher-weighted trajectory calibration, while optimizing only lightweight loop-conditioned clipping, gain, and bias parameters. Together, they act on activation distributions, precision budgets, and recursive quantization behavior to enable accurate and deployment-friendly low-bit inference for LoopLMs.
Our key contributions are summarized as follows:
\begin{itemize}
    \item We identify three key challenges in low-bit PTQ for LoopLMs: cross-loop activation dynamics, heterogeneous loop sensitivity, and the mismatch between local quantization objectives and recursive task behavior.

    \item We propose \textbf{LoopQuant}, a loop-aware PTQ framework combining \textbf{CLDC} for cross-loop distribution canonicalization, \textbf{ILPA} for interaction-aware loop-wise precision allocation, and \textbf{LCTC} for task-aligned calibration along the low-bit recursive trajectory.

    \item Extensive experiments show that LoopQuant achieves a superior accuracy--efficiency trade-off across LoopLMs, with W4A4 outperforming existing PTQ baselines at W4A8. To facilitate efficient LoopLM deployment, we will release our implementation, which, to our knowledge, is the first open-source quantization framework for LoopLMs.
\end{itemize}

\section{Related Work} 
\label{sec:related} 

\subsection{Post-Training Quantization for Large Language Models}

Post-training quantization (PTQ) compresses pretrained models without expensive end-to-end retraining, reducing the storage, memory access, and inference costs of LLMs~\citep{zhao2026specquant,zhao2026all}.
Recent advances have further explored aggressive low-bit compression and quantization-based acceleration across diverse Transformer architectures~\citep{zhao2026bwla,xu2026kbvqmoe,zhao2025quark}.
For low-bit weight--activation quantization, activation outliers remain a major obstacle to accurate quantization, particularly under stringent precision budgets~\citep{pmlr-v306-zhao26bi}.
SmoothQuant~\citep{xiao2023smoothquant} applies channel-wise scaling to shift quantization difficulty from activations to weights, while QuaRot~\citep{ashkboos2024quarot} uses orthogonal rotations to redistribute hidden-state outliers while preserving full-precision outputs.
SpinQuant~\citep{liu2025spinquant} further learns quantization-friendly rotations.
More recent methods strengthen these transformations: FlatQuant~\citep{sun2024flatquant} learns flexible affine transformations for weights and activations, while DartQuant~\citep{shao2026dartquant} performs distribution-aware rotation calibration with efficient orthogonal optimization.
Despite strong performance on standard Transformers, these methods typically calibrate each layer from local activation distributions and do not explicitly model evolving cross-loop activation dynamics or loop-dependent sensitivity of recursively shared modules.

\subsection{Looped Language Models} 

Looped Language Models (LoopLMs) increase effective computation depth by recursively reusing shared Transformer modules, decoupling model capacity from parameter count and enabling parameter-efficient scaling~\citep{dehghani2018universal}. This paradigm has been adopted by modern LLMs. Relaxed Recursive Transformers~\citep{bae2025relaxed} reuse shared Transformer layers with lightweight adapters to mitigate the expressiveness limits of strict parameter sharing. Huginn~\citep{geiping2026scaling} adopts a recurrent-depth architecture and scales test-time latent computation through recursive applications of shared modules. Ouro~\citep{zhu2026scalinglatentreasoninglooped} extends LoopLMs to large-scale pretraining, iteratively refining latent representations to improve reasoning under a fixed parameter budget. Recently, Nanbeige4.2-3B~\citep{lab2026nanbeige4} employs a Looped Transformer trained from scratch, reusing a shared layer stack to increase effective depth and showing strong performance on reasoning, coding, and agentic tasks. Low-bit LoopLM quantization has also received attention. LoopQ~\citep{fang2026loopq} provides an initial study of cross-loop distribution shifts and recursive quantization errors in Recursive Transformers. Nevertheless, evolving cross-loop activation dynamics, heterogeneous loop-wise quantization sensitivity, and error propagation along recursive trajectories remain distinct challenges for LoopLM quantization and motivate our work.

\section{Preliminaries}
\label{sec:Preliminaries}

\paragraph{Looped Language Models}

Looped Language Models (LoopLMs) increase effective computation depth by recursively reusing shared Transformer layers. Given an input sequence $\mathbf{X}$, let $\mathcal{F}_{\theta}$ denote the shared Transformer module with $L$ layers, where $\theta=\{\theta_1,\ldots,\theta_L\}$. The input is embedded as the initial hidden state $\mathbf{H}^{(0)}$, after which the shared module is recursively applied for $T$ loops:
\begin{equation}
\mathbf{H}^{(t)}
=
\mathcal{F}_{\theta}\left(\mathbf{H}^{(t-1)}\right),
\qquad
t=1,\ldots,T.
\end{equation}
The final output is obtained from the last-loop hidden state:
\begin{equation}
\mathbf{Y}
=
\operatorname{LMHead}\left(\mathbf{H}^{(T)}\right).
\end{equation}
For layer-wise notation, let $\mathbf{H}_{l}^{(t)}$ denote the output of layer $l$ at loop $t$. The computation becomes
\begin{equation}
\mathbf{H}_{l}^{(t)}
=
\mathcal{F}_{l}\left(\mathbf{H}_{l-1}^{(t)};\theta_l\right),
\qquad
\mathbf{H}_{0}^{(t)}
=
\mathbf{H}_{L}^{(t-1)}.
\end{equation}
Parameters $\theta$ are shared across loops, while hidden states evolve recursively. Thus, $L$ shared layers executed for $T$ loops provide an effective depth of approximately $LT$ while storing only $L$ layers. Despite sharing identical parameters, the layers receive activations from different recurrent states, distinguishing LoopLMs from standard feed-forward Transformers.

\section{Method}
\label{sec:method}

\begin{figure}[t]
    \centering
    \includegraphics[width=\linewidth]{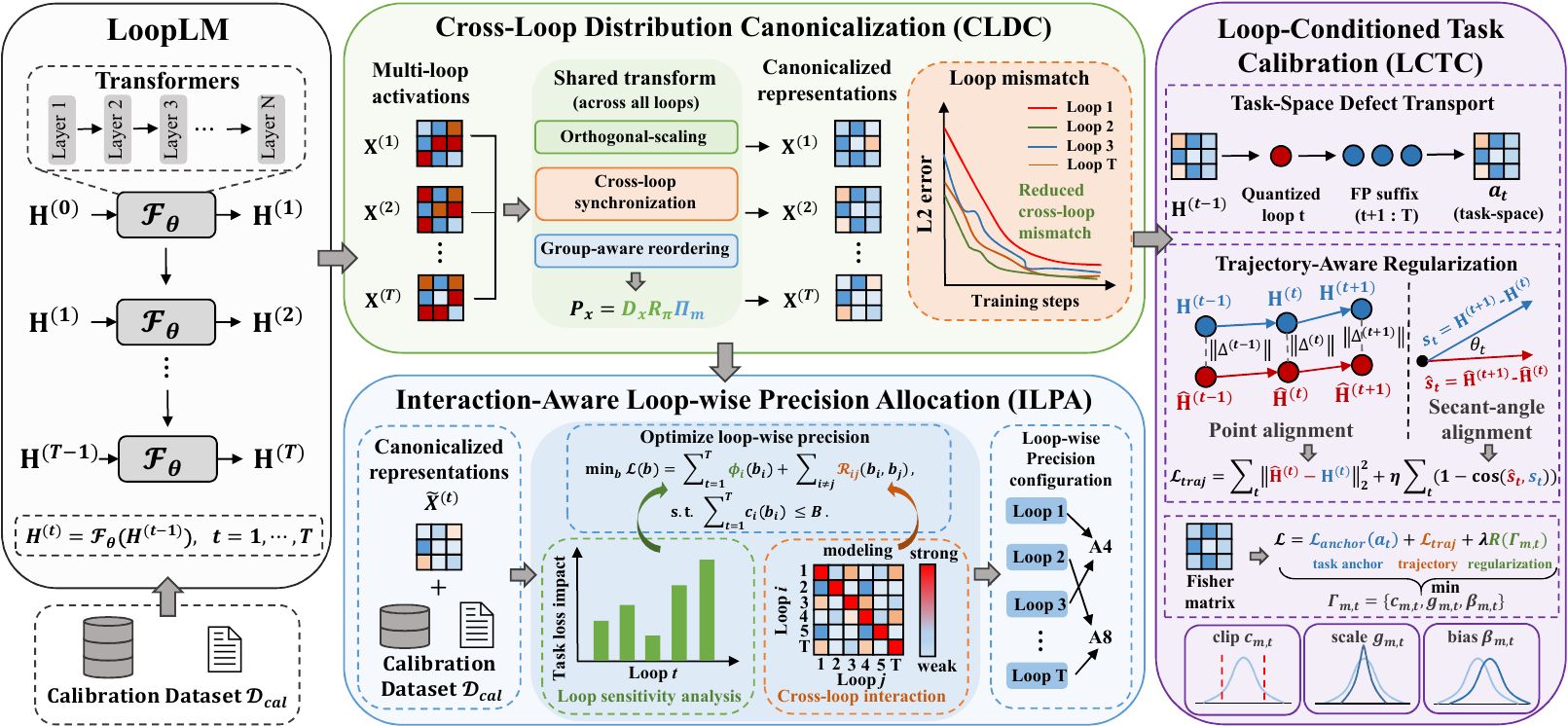}
    \caption{Overview of LoopQuant. LoopQuant consists of three stages: \textbf{CLDC} canonicalizes cross-loop activation distributions, \textbf{ILPA} allocates activation precision under a fixed bit budget, and \textbf{LCTC} aligns quantized models with recursive task behavior.}
    \label{fig:overview}
    %\vspace{-0.3cm}
\end{figure}

\paragraph{Overview.}
In this section, we first formulate low-bit quantization under recursive computation (Sec.~\ref{sec:formulate}), decomposing task degradation into loop-dependent local errors, cross-loop sensitivity and interactions, and trajectory-dependent errors.
Building on this formulation, \textbf{LoopQuant} consists of three stages (Fig.~\ref{fig:overview}): CLDC learns a shared quantization-friendly representation across recursive states (Sec.~\ref{sec:cldc}); ILPA allocates activation precision by modeling loop sensitivity and cross-loop interactions under a fixed bit budget (Sec.~\ref{sec:ilpa}); and LCTC calibrates quantization behavior along the low-bit recursive trajectory (Sec.~\ref{sec:lctc}).
Pseudocode is provided in Appendix~\ref{sec:Algorithmic}.

\subsection{Quantization under Recursive Computation}
\label{sec:formulate}

Following the LoopLM formulation in Sec.~3, we characterize low-bit quantization errors through recurrent state perturbation and the final task objective. For loop $t$, let $\mathcal{F}_{\theta}$ and $\widehat{\mathcal{F}}_{\theta}^{(t)}$ denote the full-precision and quantized mappings. Given an input state $\mathbf{H}$, define the quantization residual as
\begin{equation}
\mathbf{r}_t(\mathbf{H})
=
\widehat{\mathcal{F}}_{\theta}^{(t)}(\mathbf{H})
-
\mathcal{F}_{\theta}(\mathbf{H}).
\end{equation}

During quantized inference, loop $t$ receives $\widehat{\mathbf{H}}^{(t-1)}$, already affected by preceding quantization errors. The resulting perturbation is
\begin{equation}
\mathbf{u}_t
=
\mathbf{r}_t\left(\widehat{\mathbf{H}}^{(t-1)}\right)
=
\underbrace{
\mathbf{r}_t\left(\mathbf{H}^{(t-1)}\right)
}_{\boldsymbol{\epsilon}^{\mathrm{loc}}_t}
+
\underbrace{
\mathbf{r}_t\left(\widehat{\mathbf{H}}^{(t-1)}\right)
-
\mathbf{r}_t\left(\mathbf{H}^{(t-1)}\right)
}_{\boldsymbol{\epsilon}^{\mathrm{traj}}_t}.
\end{equation}
Here, $\boldsymbol{\epsilon}^{\mathrm{loc}}_t$ denotes the loop-dependent local error, while $\boldsymbol{\epsilon}^{\mathrm{traj}}_t$ captures the additional mismatch caused by inherited state drift. Thus,
\begin{equation}
\widehat{\mathbf{H}}^{(t)}
=
\mathcal{F}_{\theta}
\left(
\widehat{\mathbf{H}}^{(t-1)}
\right)
+
\mathbf{u}_t,
\qquad
t=1,\ldots,T.
\end{equation}

Viewing $\{\mathbf{u}_t\}_{t=1}^{T}$ as perturbations injected along the recursive trajectory, we write the task loss as
$\mathcal{L}=\mathcal{L}(\mathbf{u}_1,\ldots,\mathbf{u}_T)$.
A second-order expansion around the full-precision model gives
\begin{equation}
\Delta\mathcal{L}
\approx
\sum_{t=1}^{T}\mathbf{g}_t^{\top}\mathbf{u}_t
+
\frac{1}{2}\sum_{i=1}^{T}\sum_{j=1}^{T}
\mathbf{u}_i^{\top}\mathbf{C}_{ij}\mathbf{u}_j,
\quad
\mathbf{g}_t=
\left.\frac{\partial\mathcal{L}}{\partial\mathbf{u}_t}\right|_{\mathbf{u}=0},
\quad
\mathbf{C}_{ij}=
\left.\frac{\partial^2\mathcal{L}}{\partial\mathbf{u}_i\partial\mathbf{u}_j}\right|_{\mathbf{u}=0}.
\end{equation}
Here, $\mathbf{g}_t$ measures task sensitivity to perturbations at loop $t$, while $\mathbf{C}_{ij}$ captures cross-loop interactions. Substituting
$\mathbf{u}_t=\boldsymbol{\epsilon}^{\mathrm{loc}}_t+\boldsymbol{\epsilon}^{\mathrm{traj}}_t$
yields
\begin{equation}
\Delta\mathcal{L}
\approx\;
\sum_{t=1}^{T}
\underbrace{\mathbf{g}_t^{\top}}_{\mathrm{ILPA}}
\left(
\underbrace{\boldsymbol{\epsilon}^{\mathrm{loc}}_t}_{\mathrm{CLDC}}
+
\underbrace{\boldsymbol{\epsilon}^{\mathrm{traj}}_t}_{\mathrm{LCTC}}
\right)
+
\frac{1}{2}
\sum_{i=1}^{T}
\sum_{j=1}^{T}
\left(
\boldsymbol{\epsilon}^{\mathrm{loc}}_i
+
\boldsymbol{\epsilon}^{\mathrm{traj}}_i
\right)^{\top}
\underbrace{\mathbf{C}_{ij}}_{\mathrm{ILPA}}
\left(
\boldsymbol{\epsilon}^{\mathrm{loc}}_j
+
\boldsymbol{\epsilon}^{\mathrm{traj}}_j
\right).
\end{equation}

This decomposition links loop-dependent local error to CLDC, task sensitivity and cross-loop interaction to ILPA, and trajectory-dependent error to LCTC.

\subsection{Cross-Loop Distribution Canonicalization (CLDC)}
\label{sec:cldc}

\paragraph{Motivation.}
The shared module in LoopLMs processes recurrent states, producing loop-dependent activation ranges, outliers, and quantization-sensitive directions. Loop-specific transformations can adapt to individual states but introduce storage and cannot be fused into shared weights.

We propose \textbf{Cross-Loop Distribution Canonicalization (CLDC)}, which learns a shared quantization-friendly geometry across recurrent states. For a shared linear module with weight $\mathbf{W}_m$ and loop-specific activation $\mathbf{X}_m^{(t)}$, CLDC applies the transformation $\mathbf{P}_m$ to every loop. We parameterize its continuous component as
$\mathbf{P}_m^{\mathrm{cont}}=\mathbf{D}_m\mathbf{R}_m$,
where $\mathbf{D}_m=\operatorname{Diag}(\mathbf{s}_m)$ performs channel scaling and
$\mathbf{R}_m\mathbf{R}_m^{\top}=\mathbf{I}$ is orthogonal.
Both are optimized using calibration states from all loops, with the parameterization of $\mathbf{R}_m$ in Appendix~\ref{app:cldc_geometry}. The transformed activations and weights are
\begin{equation}
\widetilde{\mathbf{X}}_m^{(t)}
=
\mathbf{X}_m^{(t)}\mathbf{D}_m\mathbf{R}_m,
\qquad
\widetilde{\mathbf{W}}_m
=
\mathbf{W}_m\mathbf{D}_m^{-1}\mathbf{R}_m .
\end{equation}
Orthogonality preserves the full-precision computation, i.e.,
$\widetilde{\mathbf{X}}_m^{(t)}
\widetilde{\mathbf{W}}_m^{\top}
=
\mathbf{X}_m^{(t)}
\mathbf{W}_m^{\top}$ for all $t$.
Under the target quantization configuration, we define the normalized reconstruction error as
\begin{equation}
\mathcal{E}_m^{(t)}
=
\frac{
\left\|
Q_A(\widetilde{\mathbf{X}}_m^{(t)})
Q_W(\widetilde{\mathbf{W}}_m)^{\top}
-
\mathbf{X}_m^{(t)}\mathbf{W}_m^{\top}
\right\|_F^2
}{
\left\|
\mathbf{X}_m^{(t)}\mathbf{W}_m^{\top}
\right\|_F^2
}.
\end{equation}
Because severe distortion at any recurrent position can propagate forward, CLDC minimizes a smooth worst-loop objective:
\begin{equation}
\mathcal{L}_{\mathrm{rob}}
=
\tau
\log
\left(
\frac{1}{T}
\sum_{t=1}^{T}
\exp
\left(
\frac{\mathcal{E}_m^{(t)}}{\tau}
\right)
\right).
\end{equation}

Reconstruction alone does not ensure consistent group-wise quantization because outlier channels may shift across loops. CLDC therefore synchronizes cross-loop anomaly profiles. Let
$a_{m,c}^{(t)}=\operatorname{Quantile}_{q}(|\widetilde{\mathbf{X}}_{m,:,c}^{(t)}|)$,
$p_{m,c}^{(t)}=a_{m,c}^{(t)}/\sum_k a_{m,k}^{(t)}$, and
$\overline{\mathbf{p}}_m=\frac{1}{T}\sum_t\mathbf{p}_m^{(t)}$.
The synchronization loss is
\begin{equation}
\mathcal{L}_{\mathrm{sync}}
=
\frac{1}{T}
\sum_{t=1}^{T}
\|
\mathbf{p}_m^{(t)}
-
\overline{\mathbf{p}}_m
\|_2^2 .
\end{equation}
To avoid excessive scaling, we further use
$\mathcal{L}_{\mathrm{reg}}
=
\frac{1}{d_i}\|\mathbf{s}_m-\mathbf{1}\|_2^2$,
giving
\begin{equation}
(\mathbf{D}_m^\star,\mathbf{R}_m^\star)
=
\arg\min
\left[
\mathcal{L}_{\mathrm{rob}}
+
\lambda_{\mathrm{sync}}\mathcal{L}_{\mathrm{sync}}
+
\lambda_{\mathrm{reg}}\mathcal{L}_{\mathrm{reg}}
\right].
\end{equation}

Finally, CLDC learns a hard channel permutation $\mathbf{\Pi}_m^\star$ to improve quantization compatibility by grouping channels with similar cross-loop dynamic ranges; the grouping objective and optimization are described in Appendix~\ref{app:cldc_reorder}. The final transformation is
$\mathbf{P}_m^\star=\mathbf{D}_m^\star\mathbf{R}_m^\star\mathbf{\Pi}_m^\star$.
The resulting representation is frozen for subsequent precision allocation and task calibration.

\subsection{Interaction-Aware Loop-wise Precision Allocation (ILPA)}
\label{sec:ilpa}

\paragraph{Motivation.}
After CLDC, loops may still require different activation precisions because their perturbations have unequal task impact and interact through recursive propagation. Uniform precision ignores this heterogeneity, while local reconstruction errors cannot capture cross-loop coupling.

Inspired by cooperative optimization for mixed-precision quantization~\citep{zhao2025coopqcooperativegameinspired}, we propose \textbf{Interaction-Aware Loop-wise Precision Allocation (ILPA)}, which assigns loop-wise activation precision from complete recursive trajectories under a fixed bit budget. Let $\mathcal{T}=\{1,\ldots,T\}$ and $\mathcal{S}\subseteq\mathcal{T}$ denote loops using low precision $b_{\mathrm L}$, while the others use $b_{\mathrm H}$. For each configuration, ILPA keeps weight quantization and model parameters fixed and evaluates the token-level NLL:
\begin{equation}
v(\mathcal{S})
=
\mathbb{E}_{\mathbf{x}\sim\mathcal{D}_{\mathrm{cal}}}
\left[
-\frac{1}{|\mathbf{x}|}
\sum_n
\log p_{\mathcal{S}}(x_{n+1}|x_{\leq n})
\right].
\end{equation}
Since $v(\mathcal{S})$ is evaluated over the complete low-bit trajectory, it captures loop-specific functionality, error propagation, and joint precision effects.
To estimate loop sensitivity, ILPA aggregates context-dependent marginal degradation following the Shapley principle~\citep{shapley1953value}:
\begin{equation}
\phi_t
=
\frac{1}{T!}
\sum_{\pi\in\mathfrak{S}_T}
\Delta_t^{(\pi)} .
\end{equation}
The resulting $\phi_t$ provides a discrete task-level estimate of the
first-order sensitivity induced by reducing the precision of loop $t$;
the marginal construction is provided in Appendix~\ref{app:ilpa_sensitivity}.
ILPA further models pairwise recursive interaction:
\begin{equation}
\Gamma_{ij}
=
\frac{1}{T!}
\sum_{\pi\in\mathfrak{S}_T}
\left(
\Delta_{ij}^{(\pi)}
-
\Delta_i^{(\pi)}
-
\Delta_j^{(\pi)}
\right).
\end{equation}
$\Gamma_{ij}$ indicates stronger-than-additive degradation when loops $i$ and $j$ are jointly assigned low precision. We normalize $\Gamma_{ij}$ across precision contexts to obtain $\widetilde{\Gamma}_{ij}$, which captures the discrete task-level interaction effect induced by joint precision reduction; details are given in Appendix~\ref{app:ilpa_interaction}.
Given target activation bit-width $\overline b$, ILPA minimizes a discrete surrogate objective:
\begin{equation}
\mathcal{J}(\mathbf q)
=
\sum_{t=1}^{T}
q_t\phi_t
+
\lambda
\sum_{i<j}
q_iq_j\widetilde{\Gamma}_{ij},
\end{equation}
where $q_t=1$ indicates low precision. The first term captures individual loop sensitivity, while the second accounts for recursive interaction. For small $T$, ILPA solves the constrained assignment by exact enumeration; for larger models, a greedy solver selects precision reductions with the smallest marginal increase in $\mathcal{J}$ until the bit budget is met. Solver and complexity details are provided in Appendix~\ref{app:ilpa_solver}. The resulting assignment remains fixed during inference.

\subsection{Loop-Conditioned Task Calibration (LCTC)}
\label{sec:lctc}

\paragraph{Motivation.}
After CLDC and ILPA, a loop with small local quantization error may behave
differently on states shifted by preceding low-bit computation, producing
trajectory-dependent error $\boldsymbol{\epsilon}^{\mathrm{traj}}_t$.
Local reconstruction alone therefore cannot ensure robust behavior along the
recursive trajectory.

We propose \textbf{Loop-Conditioned Task Calibration (LCTC)}, which
calibrates loop-dependent quantization responses while keeping the
quantization structure and precision assignment fixed. For module $m$ at loop
$t$, LCTC introduces static parameters
$\boldsymbol{\Gamma}_{m,t}
=
\{\mathbf c_{m,t},\mathbf g_{m,t},\boldsymbol{\beta}_{m,t}\}$,
controlling group-wise clipping, gain, and bias. Given fixed quantized weight
$\widehat{\mathbf W}_m$, the calibrated mapping is
\begin{equation}
\widehat{\mathbf y}_{m,t}
=
\mathbf g_{m,t}
\odot
\left[
\widehat{\mathbf W}_m
Q_{b_t}
(
\mathbf x_{m,t};
\mathbf c_{m,t}
)
+
\mathbf b_m
\right]
+
\boldsymbol{\sigma}_{m,t}
\odot
\boldsymbol{\beta}_{m,t},
\end{equation}
where $\boldsymbol{\sigma}_{m,t}$ denotes the fixed group-wise output
normalization statistic. These parameters depend only on module and loop
index and are stored statically, requiring no token-wise history or online
optimization, which keeps deployment lightweight without extra dynamic state.
To align quantization defects with task behavior, let
$\mathcal S_{t+1}(\cdot)$ map the output of loop $t$ through the remaining
computation to the final logits. We define the task-space defect as
\begin{equation}
\mathbf d_t
(
\mathbf h;\boldsymbol{\Gamma}
)
=
\mathcal S_{t+1}
\left(
\widehat{\mathcal F}_{t,\boldsymbol{\Gamma}}
(\mathbf h)
\right)
-
\mathcal S_t(\mathbf h).
\end{equation}
For the full-precision state $\mathbf H^{(t-1)}$, let
$\mathbf a_t=
\mathbf d_t(\mathbf H^{(t-1)};\boldsymbol{\Gamma})$.
The low-bit trajectory inherits drift
$\boldsymbol{\delta}_{t-1}
=
\widehat{\mathbf H}^{(t-1)}
-
\mathbf H^{(t-1)}$.
LCTC measures its effect on the defect through the secant

\begin{equation}
\mathbf s_t(\alpha)
=
\frac{
\mathbf d_t
(
\mathbf H^{(t-1)}
+
\alpha\boldsymbol{\delta}_{t-1};
\boldsymbol{\Gamma}
)
-
\mathbf d_t
(
\mathbf H^{(t-1)};
\boldsymbol{\Gamma}
)
}
{\alpha}.
\end{equation}
This term provides a task-space approximation of the trajectory-dependent
error, with derivation given in Appendix~\ref{app:lctc_transport}.
Because logit perturbations have unequal effects on predictions, LCTC uses
the categorical Fisher metric
$\mathbf F(\mathbf p)=\operatorname{Diag}(\mathbf p)-\mathbf p\mathbf p^{\top}$,
where $\mathbf p=\operatorname{softmax}(\mathbf z)$, with Fisher energy
$\mathcal E_{\mathrm F}(\mathbf v)=
\mathbf v^{\top}\mathbf F(\mathbf p)\mathbf v$.
We optimize the clean-state defect and trajectory sensitivity as
\begin{equation}
\mathcal L_{\mathrm{anchor}}
=
\mathbb E_{t,\mathbf x}
[
\mathcal E_{\mathrm F}(\mathbf a_t)
],
\qquad
\mathcal L_{\mathrm{traj}}
=
\mathbb E_{t,\mathbf x,\alpha}
[
\mathcal E_{\mathrm F}(\mathbf s_t(\alpha))
].
\end{equation}
The objective combines teacher alignment, language modeling, and
trajectory-aware correction:
\begin{equation}
\mathcal L_{\mathrm{LCTC}}
=
\mathcal L_{\mathrm{KD}}
+
\lambda_{\mathrm{NLL}}\mathcal L_{\mathrm{NLL}}
+
\lambda_{\mathrm a}\widetilde{\mathcal L}_{\mathrm{anchor}}
+
\lambda_{\mathrm t}\widetilde{\mathcal L}_{\mathrm{traj}}
+
\lambda_{\mathrm{reg}}
\mathcal R(\boldsymbol{\Gamma}).
\end{equation}
Normalization and regularization details are provided in
Appendix~\ref{app:lctc_fisher}. Only
$\boldsymbol{\Gamma}_{m,t}$ is optimized, while quantized weights, CLDC
transformations, ILPA assignments, and parameters remain frozen. Thus, LCTC
performs task-aligned calibration rather than retraining the quantized model.

\section{Experiments}
\label{sec:Experiments}
\subsection{Experiment setup}

\paragraph{Models and Datasets.}
We evaluate \textbf{LoopQuant} on representative LoopLMs, including Ouro (with thinking-enabled models)~\citep{zhu2026scalinglatentreasoninglooped}, Huginn~\citep{geiping2026scaling}, and Nanbeige4.2-3B~\citep{lab2026nanbeige4}. 
We report WikiText2 and C4 perplexity (PPL)~\citep{wikitext2, c4}, together with seven zero-shot tasks: ARC-Challenge/Easy~\citep{clark2018thinksolvedquestionanswering}, HellaSwag~\citep{zellers2019hellaswagmachinereallyfinish}, LAMBADA-openai/standard~\citep{paperno2016lambadadatasetwordprediction}, PIQA~\citep{bisk2019piqareasoningphysicalcommonsense}, and WinoGrande~\citep{sakaguchi2019winograndeadversarialwinogradschema}. 
We further evaluate on reasoning and generation benchmarks, including MMLU~\citep{mmlu}, GSM8K~\citep{gsm8k}, and HumanEval~\citep{humaneval}.

\paragraph{Baseline Methods.}
We compare LoopQuant with representative low-bit PTQ methods, including SmoothQuant~\citep{xiao2023smoothquant} with channel scaling, QuaRot~\citep{ashkboos2024quarot} with orthogonal rotation, SpinQuant~\citep{liu2025spinquant} with learnable rotation, FlatQuant~\citep{sun2024flatquant} with affine transformation, and DartQuant~\citep{shao2026dartquant} with distribution-aware rotation calibration. 
BF16 serves as the full-precision reference. 
We further compare with LoopQ~\citep{fang2026loopq}; since its code is not publicly released, we report the results in its original paper in Appendix~\ref{app:loopq_comparison}.

\paragraph{Implementation Details.}
We use per-channel symmetric quantization for weights and per-token symmetric quantization for activations. 
For fair comparison, all methods use GPTQ~\citep{frantar2022gptq} as the weight quantization backend and 128 sampled WikiText2 sequences of length 1024 for calibration (Appendix~\ref{app:calibration_ablation}). 
CLDC optimizes loop-shared parameters with Adam (Appendix~\ref{app:cldc_optimization}), and LCTC details are provided in Appendix~\ref{app:lctc_optimization}. ILPA uses exact enumeration for fewer than 5 loops and 100 Monte Carlo permutations otherwise.
All experiments are conducted on NVIDIA RTX PRO 6000 GPUs, and accuracy is evaluated with lm-eval~\citep{gao2024lmeval}.

\subsection{Main Results}

\paragraph{Comparison Results.}

\begin{table*}[t!]
\centering
%\vspace{-0.3cm}

\caption{
Comparison of LoopQuant with representative low-bit PTQ methods on LoopLMs.
Bold and underline denote the \textbf{best} and \underline{second-best}
quantized results.
}

\vspace{-0.2cm}

\resizebox{0.9\textwidth}{!}{%
\tablestyle{2.5pt}{1.2}
\begin{tabular}{l|c|ccc|ccc|ccc|ccc}
\toprule

\multirow{2}{*}{\textbf{Method}}
&
\multirow{2}{*}{\textbf{W/A}}
&
\multicolumn{3}{c}{\textbf{Ouro-1.4B}}
&
\multicolumn{3}{c}{\textbf{Ouro-2.6B}}
&
\multicolumn{3}{c}{\textbf{Nanbeige4.2-3B}}
&
\multicolumn{3}{c}{\textbf{Huginn-0125}}
\\

\cdashline{3-14}

&
&
Avg.$\uparrow$
&
WT2$\downarrow$
&
C4$\downarrow$
&
Avg.$\uparrow$
&
WT2$\downarrow$
&
C4$\downarrow$
&
Avg.$\uparrow$
&
WT2$\downarrow$
&
C4$\downarrow$
&
Avg.$\uparrow$
&
WT2$\downarrow$
&
C4$\downarrow$
\\

\hdashline

BF16
&
16/16
&
66.85&8.60&12.61
&
71.12&7.61&11.42
&
58.68&24.17&31.45
&
49.89&12.10&16.28
\\

\hdashline

SmoothQuant
&
4/8
&
60.65&11.25&15.38
&
67.12&9.48&13.46
&
\underline{56.10}&28.63&\underline{32.38}
&
46.95&13.41&17.65
\\

QuaRot
&
4/8
&
62.53&10.78&15.21
&
\underline{68.30}&\underline{9.09}&\underline{13.02}
&
53.83&34.09&41.54
&
\underline{47.73}&13.38&17.63
\\

SpinQuant
&
4/8
&
\underline{62.69}&10.79&15.06
&
67.00&9.11&13.03
&
54.47&31.10&35.86
&
47.33&\underline{13.34}&\underline{17.62}
\\

FlatQuant
&
4/8
&
61.37&\underline{10.65}&\underline{14.84}
&
67.31&9.26&13.18
&
55.14&\underline{25.99}&32.93
&
47.31&13.45&17.71
\\

DartQuant
&
4/8
&
60.52&11.45&15.63
&
65.65&10.04&14.01
&
53.06&30.69&38.15
&
47.32&13.50&17.76
\\

\rowcolor{gray!5}
\textbf{LoopQuant}
&
4/6
&
\textbf{64.62}&\textbf{9.48}&\textbf{13.57}
&
\textbf{70.20}&\textbf{8.04}&\textbf{11.94}
&
\textbf{56.74}&\textbf{25.84}&\textbf{32.19}
&
\textbf{49.09}&\textbf{12.47}&\textbf{16.73}
\\

\hdashline

SmoothQuant
&
4/4
&
54.29&13.99&18.36
&
61.76&11.41&15.54
&
\underline{50.70}&33.85&38.82
&
44.10&14.92&19.30
\\

QuaRot
&
4/4
&
\underline{58.13}&\underline{12.50}&17.15
&
\underline{63.58}&\underline{10.90}&\underline{14.90}
&
49.74&34.72&41.88
&
44.53&\underline{14.76}&\underline{19.07}
\\

SpinQuant
&
4/4
&
55.77&13.54&18.43
&
58.94&13.23&18.00
&
48.33&33.91&37.95
&
44.50&14.95&19.35
\\

FlatQuant
&
4/4
&
57.14&12.71&\underline{17.04}
&
62.57&10.96&15.11
&
49.14&\underline{30.44}&\underline{36.39}
&
\underline{44.56}&14.98&19.37
\\

DartQuant
&
4/4
&
51.46&15.77&20.49
&
55.69&14.37&18.67
&
46.19&36.28&38.76
&
44.06&15.17&19.57
\\

\rowcolor{gray!5}
\textbf{LoopQuant}
&
4/4
&
\textbf{64.04}&\textbf{9.86}&\textbf{14.00}
&
\textbf{68.89}&\textbf{8.27}&\textbf{12.24}
&
\textbf{56.11}&\textbf{26.64}&\textbf{33.02}
&
\textbf{48.39}&\textbf{12.78}&\textbf{17.11}
\\

\bottomrule
\end{tabular}%
}

\vspace{-0.5cm}

\label{tab:main_results}

\end{table*}
As shown in Table~\ref{tab:main_results}, LoopQuant consistently outperforms low-bit PTQ methods across LoopLM architectures under $W4A6$ and $W4A4$. 
Here, $W4A6$ uses loop-wise mixed activation precision: half the loops use 4-bit activations and half use 8-bit, averaging 6 bits. 
Despite lower activation precision than $W4A8$ baselines, LoopQuant achieves superior performance across all models. 
For example, on Ouro-2.6B, LoopQuant achieves 70.20\% average zero-shot accuracy under $W4A6$, outperforming the strongest baseline by 1.90 percentage points (pp) while reducing WikiText2 PPL from 9.09 to 8.04. 
Under $W4A4$, existing methods suffer substantial degradation, whereas LoopQuant maintains performance. 
On Huginn-0125 ($T=6$), LoopQuant achieves 48.39\% average zero-shot accuracy, substantially outperforming the best baseline (44.56\%). 
These results show that LoopQuant mitigates quantization error propagation and provides a favorable accuracy--compression trade-off across LoopLMs.
Full results are in Appendix~\ref{app:detailed main results}.

\paragraph{Experiments on Reasoning-oriented Models.}
Thinking-oriented LLMs rely on extended reasoning and complex intermediate computation, making them highly sensitive to quantization errors. 
We further evaluate LoopQuant on Ouro-1.4B-thinking and Ouro-2.6B-thinking using MMLU, GSM8K, and HumanEval (Table~\ref{tab:reasoning_results}). 
Under $W4A6$, LoopQuant outperforms all $W4A8$ baselines 
\begin{wraptable}{r}{0.62\textwidth}
    \centering
    %\vspace{-0.3cm}
    \caption{Evaluation on reasoning-oriented LoopLMs.}
    %\vspace{-0.3cm}
    \resizebox{\linewidth}{!}{
        \tablestyle{2.5pt}{1.2}
\begin{tabular}{l|c|ccc|ccc}
\toprule

\multirow{2}{*}{\textbf{Method}}
&
\multirow{2}{*}{\textbf{W/A}}
&
\multicolumn{3}{c|}{\textbf{Ouro-1.4B-thinking}}
&
\multicolumn{3}{c}{\textbf{Ouro-2.6B-thinking}}
\\

\cdashline{3-8}

&
&
MMLU
&
GSM8K
&
HumanEval
&
MMLU
&
GSM8K
&
HumanEval
\\

\hdashline

BF16
&
16/16
&
67.71&69.57&71.34
&
76.20&73.31&82.32
\\

\hdashline

SmoothQuant
&
4/8
&
59.42&34.38&\underline{63.24}
&
68.97&60.56&74.22
\\

QuaRot
&
4/8
&
57.24&34.23&58.98
&
\underline{71.50}&\underline{65.94}&\underline{79.71}
\\

SpinQuant
&
4/8
&
56.68&53.25&60.80
&
71.26&63.67&75.44
\\

FlatQuant
&
4/8
&
\underline{60.50}&\underline{59.09}&61.02
&
70.21&63.59&77.27
\\

DartQuant
&
4/8
&
57.71&53.25&58.98
&
67.19&63.75&71.17
\\
\rowcolor{gray!5}
\textbf{LoopQuant}
&
4/6
&
\textbf{61.20}&\textbf{61.49}&\textbf{70.73}
&
\textbf{72.74}&\textbf{72.84}&\textbf{81.71}
\\

\hdashline

SmoothQuant
&
4/4
&
50.54&11.83&41.34
&
62.46&43.69&61.85
\\

QuaRot
&
4/4
&
51.86&29.42&\underline{49.66}
&
\underline{67.09}&\underline{58.70}&\underline{71.00}
\\

SpinQuant
&
4/4
&
44.78&18.20&14.63
&
63.67&43.69&64.90
\\

FlatQuant
&
4/4
&
\underline{52.59}&\underline{41.46}&52.32
&
64.86&57.26&61.85
\\

DartQuant
&
4/4
&
41.99&6.52&20.73
&
50.76&20.92&4.88
\\
\rowcolor{gray!5}
\textbf{LoopQuant}
&
4/4
&
\textbf{59.51}&\textbf{60.88}&\textbf{60.98}
&
\textbf{72.46}&\textbf{72.02}&\textbf{81.15}
\\

\bottomrule
\end{tabular}

\vspace{-0.3cm}

\label{tab:reasoning_results}

    }
    %\vspace{-0.3cm}
\end{wraptable}
on both models. For example, on Ouro-1.4B-thinking, LoopQuant achieves 61.49\% GSM8K and 70.73\% HumanEval accuracy, outperforming the strongest baseline by 7.49 pp on HumanEval. 
At $W4A4$, existing methods suffer significant degradation on mathematical and code reasoning tasks, with some nearly collapsing. 
In contrast, LoopQuant retains over 97\% of BF16 performance averaged across MMLU, GSM8K, and HumanEval on Ouro-2.6B-thinking. 
These results show that LoopQuant mitigates quantization-induced error accumulation in recursive reasoning (Appendix~\ref{app:layerwise_error}) while preserving high-level reasoning under extremely low-bit inference.

\subsection{Ablation Studies}

\paragraph{Component Ablation.}
We conduct an ablation on Nanbeige4.2-3B under $W4A6$ to evaluate each LoopQuant component. 
As shown in Table~\ref{tab:ablation_loopquant}, each component improves quantization performance, addressing different sources of recursive inference error. 
Starting from the vanilla $W4A6$ baseline, CLDC reduces WikiText2 PPL from 47.39 to 28.41 by regularizing activation distributions (Appendix~\ref{app:cldc_distribution}). 
Adding ILPA further reduces WikiText2 and C4 PPL to 27.05 and 34.17, respectively, by allocating activation precision according to loop-specific sensitivity. 
\begin{wraptable}{r}{0.6\textwidth}
    \centering
    %\vspace{-0.3cm}
    \caption{Impact of different components in LoopQuant.}
    %\vspace{-0.3cm}
    \resizebox{\linewidth}{!}{
        \tablestyle{2.5pt}{1.2}
\begin{tabular}{c | c:c:c | c | c | c | c | c}
\toprule
\multirow{2}{*}{\textbf{W/A}} 
& \multicolumn{3}{c|}{\textbf{CLDC}} 
& \multirow{2}{*}{\textbf{ILPA}} 
& \multirow{2}{*}{\textbf{LCTC}} 
& \multirow{2}{*}{\textbf{Storage Overhead}} 
& \textbf{Wiki2} 
& \textbf{C4} \\
& $\mathbf{D}_m^{\star}$ & $\mathbf{R}_m^{\star}$ & $\mathbf{\Pi}_m^{\star}$ & & & &\textbf{PPL}($\downarrow$) &\textbf{PPL}($\downarrow$) \\
\midrule

\multirow{6}{*}{W4A6}
& \xmark & \xmark & \xmark 
& \multirow{4}{*}{\LARGE\xmark} 
& \multirow{4}{*}{\LARGE\xmark} 
& 0 
& 47.39 
& 53.65 \\

& \cmark & \xmark & \xmark 
&  
&  
& +1.42 MB (0.04\%) 
& 36.91 
& 41.52 \\

& \cmark & \cmark & \xmark 
&  
&  
& +46.84 MB (1.22\%) 
& 28.41 
& 36.31 \\

& \cmark & \cmark & \cmark 
&  
&  
& +52.51 MB (1.37\%) 
& 27.85 
& 35.10 \\

\cdashline{2-9}

& \multicolumn{3}{c|}{\multirow{2}{*}{\LARGE\cmark}}
& \cmark 
& \xmark 
& +52.51 MB (1.37\%) 
& 27.05 
& 34.17 \\
% \rowcolor{gray!5}
& &  &  
& \cmark 
& \cmark 
& +53.08 MB (1.39\%) 
& \textbf{25.84} 
& \textbf{32.19} \\

\bottomrule
\end{tabular}
\label{tab:ablation_loopquant}
    }
    %\vspace{-0.4cm}
\end{wraptable}
With LCTC, LoopQuant achieves 25.84 WikiText2 and 32.19 C4 PPL, demonstrating the importance of trajectory-aware correction for accumulated quantization errors (Appendix~\ref{app:lctc_parameter_ablation}).
Moreover, the framework adds only 53.08 MB storage, corresponding to 1.39\% of the model size, indicating negligible memory cost. 
These results confirm that CLDC, ILPA, and LCTC are complementary, enabling effective and efficient low-bit quantization for recursive language models.

\paragraph{Ablation of Loop-wise Bit Allocation.}
We further analyze ILPA by comparing different loop 
\begin{wraptable}{r}{0.59\textwidth}
    \centering
    %\vspace{-0.3cm}
    \caption{Comparison of loop-wise bit allocation strategies.}
    %\vspace{-0.3cm}
    \resizebox{\linewidth}{!}{
        \tablestyle{2.5pt}{1.2}
\begin{tabular}{c|c|c|c|c}
\toprule
\multirow{2}{*}{\textbf{Model}}
&
\textbf{Allocation}
&
\textbf{Loop-wise Bits}
&
\textbf{Wiki2}
&
\textbf{Avg.}
\\
&\textbf{Strategy} & ($t=1\rightarrow6$) &\textbf{PPL}($\downarrow$) &\textbf{Acc.}$\uparrow$ \\
\midrule

\multirow{5}{*}{\makecell{Huginn-0125\\($T=6$)}}
&
Uniform
&
[6,6,6,6,6,6]
&
13.82
&
46.92
\\

&
Random
&
3 $\times$ A4 + 3 $\times$ A8
&
15.21
&
44.83
\\

&
Local Sensitivity
&
[4,8,4,8,8,4]
&
13.05
&
48.01
\\

&
Marginal Sensitivity
&
[4,8,8,4,4,8]
&
13.67
&
47.54
\\
\rowcolor{gray!5}
&
\textbf{ILPA}
&
\textbf{[4,4,4,8,8,8]}
&
\textbf{12.47}
&
\textbf{49.09}
\\

\bottomrule
\end{tabular}

\label{tab:ilpa_ablation}
    }
    %\vspace{-0.4cm}
\end{wraptable}
-wise activation bit allocation strategies on Huginn-0125 ($T=6$). 
As shown in Table~\ref{tab:ilpa_ablation}, all strategies maintain the same average activation precision of 6 bits. 
Compared with uniform allocation, random allocation (averaged over five sampled patterns) may assign low precision to sensitive loops, degrading performance. 
Although local and marginal sensitivity-based heuristics achieve better results, they still underperform ILPA, indicating that sensitivity alone cannot capture the optimal precision assignment. 
By globally optimizing loop-wise precision allocation, ILPA achieves 12.47 WikiText2 perplexity and 49.09\% average zero-shot accuracy. Further validation of ILPA across different quantization backbones is
provided in Appendix~\ref{app:ilpa_backbone}.

\subsection{Efficiency Analysis of LoopQuant}

\begin{figure}[t]
    \centering
    \includegraphics[width=0.95\linewidth]{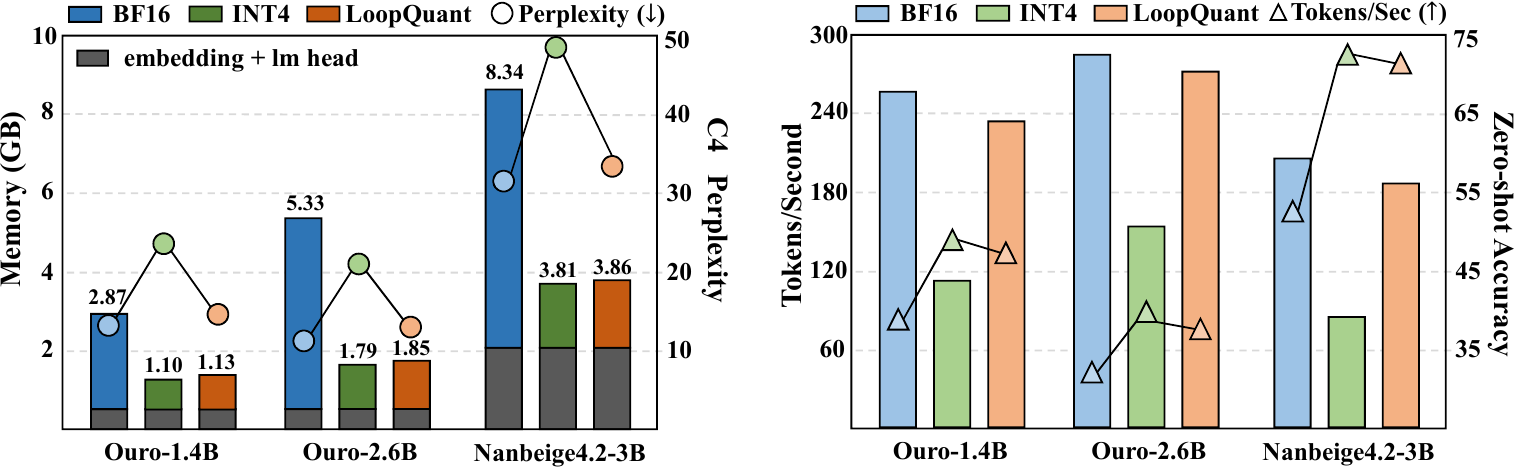}
    %\vspace{-0.2cm}
    \caption{Efficiency analysis. \textbf{Left}: Memory consumption and C4 perplexity of BF16, INT4, and LoopQuant across three LoopLMs, where gray regions denote the embedding and LM head. \textbf{Right}: Decoding throughput (Tokens/Sec) and average zero-shot accuracy under the same settings.}
    \label{fig:efficiency}
    %\vspace{-0.4cm}
\end{figure}

\paragraph{Memory Savings.}
We evaluate the storage efficiency of LoopQuant against BF16 and naive INT4 RTN across three LoopLMs, as shown in Fig.~\ref{fig:efficiency} (left).
LoopQuant reduces overall model memory by 53.7--65.3\% compared with BF16, while the quantized linear layers alone achieve approximately 74\% memory savings.
Compared with naive INT4 RTN, LoopQuant introduces at most 3.4\% additional memory overhead, largely preserving the storage efficiency of standard 4-bit quantization (Appendix~\ref{app:storage_overhead}).
Despite this negligible overhead, LoopQuant substantially mitigates C4 perplexity degradation caused by naive INT4 and achieves performance close to BF16.

\paragraph{Inference Speedup.}
We evaluate LoopQuant inference efficiency with PyTorch on NVIDIA RTX PRO 6000 GPUs, using one GPU per model.
With batch size 4, we measure decoding throughput by generating 1,024 tokens after prefilling 4,096 tokens.
As shown in Fig.~\ref{fig:efficiency} (right), LoopQuant improves throughput by 1.5--1.7$\times$ over BF16 across three models.
Although naive INT4 attains slightly higher throughput, it suffers severe accuracy degradation; LoopQuant preserves most INT4 efficiency gain while maintaining zero-shot accuracy close to BF16.

\section{Conclusion}
\label{sec:conclusion}
We present LoopQuant, a loop-aware PTQ framework for low-bit LoopLM quantization.
LoopQuant addresses cross-loop activation dynamics, heterogeneous loop sensitivity, and misalignment between local objectives and recursive task behavior.
CLDC learns a shared quantization-friendly representation, ILPA allocates loop-wise activation precision under a fixed bit budget, and LCTC calibrates quantization along the low-bit recursive trajectory.
Across LoopLMs, LoopQuant consistently outperforms low-bit PTQ methods; W4A4 matches or surpasses W4A8 baselines while preserving reasoning capability.
It delivers substantial memory savings and decoding speedup with negligible storage overhead, enabling accurate and deployment-friendly low-bit LoopLM inference.

\clearpage

\subsection*{AI use statement}

Generative AI tools were used during manuscript preparation only for language-related assistance, including proofreading, translation, concision, and improvements to readability. They were not used to generate the core research ideas, methodological contributions, experimental protocols, or scientific conclusions. Any AI-supported edits were manually inspected and, when necessary, revised by the authors before inclusion in the manuscript. Responsibility for the technical content, reported results, and final presentation rests entirely with the authors.

\subsection*{Ethics statement}

This work studies LoopQuant, a post-training quantization framework for improving the deployment efficiency of Looped Language Models. Our experiments are conducted on existing pretrained models and publicly available calibration and evaluation benchmarks. Reducing memory and inference costs may broaden access to LoopLMs, but greater deployment efficiency can also facilitate unintended or harmful uses. Quantized models may inherit biases, factual inaccuracies, and unsafe behaviors from their full-precision counterparts, and low-bit quantization may alter model behavior in ways not captured by standard accuracy benchmarks. Our experiments evaluate predictive performance and system efficiency rather than safety, fairness, privacy, or robustness guarantees. We therefore recommend application-specific evaluation and appropriate safeguards before real-world deployment, together with compliance with the licenses and usage policies of the underlying models and datasets.

\subsection*{Reproducibility statement}

To facilitate reproducibility, we present the recursive quantization formulation and the three components of LoopQuant in Secs.~\ref{sec:formulate}--\ref{sec:lctc}, with detailed pseudocode, derivations, and implementation details provided in Appendices~\ref{sec:Algorithmic} and~\ref{sec:proof}. Sec.~\ref{sec:Experiments} specifies the evaluated models, datasets, calibration procedure, quantization settings, evaluation protocol, and hardware. Detailed task-level results and calibration robustness analyses are reported in Appendices~\ref{app:detailed main results} and~\ref{app:calibration_ablation}, while deployment storage accounting is provided in Appendix~\ref{app:storage_overhead}. We also provide a public repository containing the LoopQuant implementation, quantization and evaluation scripts, and experimental configurations to support reproduction of the reported results.

\bibliography{iclr2027_conference}
\bibliographystyle{iclr2027_conference}

\clearpage
\appendix

\section{Appendix}
\label{sec:appendix}

\section*{Appendix Overview}
This appendix is organized as follows.
\begin{itemize}
    \item Section~\ref{sec:Algorithmic}: Pseudocode of LoopQuant.
    \begin{itemize}
        \item Section~\ref{sec:code-loopquant}: Overall Workflow of LoopQuant.
        \item Section~\ref{sec:code-cldc}: Cross-Loop Distribution Canonicalization (CLDC).
        \item Section~\ref{sec:code-ilpa}: Interaction-Aware Loop-wise Precision Allocation (ILPA).
        \item Section~\ref{sec:code-lctc}: Loop-Conditioned Task Calibration (LCTC).
    \end{itemize}

    \item Section~\ref{sec:proof}: Detailed Derivations and Implementation Details.
    \begin{itemize}
        \item Section~\ref{app:cldc_geometry}: Orthogonal Canonical Geometry Parameterization.
        \item Section~\ref{app:cldc_sync}: Robust Cross-Loop Synchronization Objective.
        \item Section~\ref{app:cldc_reorder}: Group-Aware Channel Reordering.
        \item Section~\ref{app:ilpa_sensitivity}: Shapley-Style Loop Sensitivity Estimation.
        \item Section~\ref{app:ilpa_interaction}: Recursive Interaction Modeling.
        \item Section~\ref{app:ilpa_solver}: Exact and Greedy Precision Allocation.
        \item Section~\ref{app:lctc_param}: Loop-Conditioned Calibration Parameterization.
        \item Section~\ref{app:lctc_transport}: Task-Space Defect Transport.
        \item Section~\ref{app:lctc_fisher}: Fisher-Weighted Trajectory Calibration.
    \end{itemize}

    \item Section~\ref{app:experimental_details}: Additional Experimental Details.
    \begin{itemize}
        \item Section~\ref{app:cldc_optimization}: CLDC Optimization Details.
        \item Section~\ref{app:lctc_optimization}: LCTC Optimization Details.
        \item Section~\ref{app:storage_overhead}: Deployment Storage Overhead.
        \item Section~\ref{app:detailed main results}: Detailed Main Results.
        \item Section~\ref{app:loopq_comparison}: Comparison with LoopQ.
        \item Section~\ref{app:calibration_ablation}: Ablation on Calibration Data.
        \item Section~\ref{app:ilpa_backbone}: Effectiveness of ILPA with Different Quantization Backbones.
        \item Section~\ref{app:lctc_parameter_ablation}: Ablation on LCTC Calibration Parameters.
        \item Section~\ref{app:layerwise_error}: Layer-wise Quantization Error during Reasoning.
        \item Section~\ref{app:cldc_distribution}: Activation Distribution Transformation by CLDC.
    \end{itemize}
\end{itemize}

\subsection{Pseudocode of LoopQuant}
\label{sec:Algorithmic}

This section provides the detailed pseudocode of LoopQuant.
The complete quantization procedure consists of three stages:
Cross-Loop Distribution Canonicalization (CLDC),
Interaction-Aware Loop-wise Precision Allocation (ILPA),
and Loop-Conditioned Task Calibration (LCTC).
CLDC learns a shared quantization-friendly representation across recursive
states, ILPA determines loop-wise activation precision under a fixed bit-width
budget, and LCTC further aligns the quantized model with recursive task behavior
through task-space calibration.

\subsubsection{Overall Workflow of LoopQuant}
\label{sec:code-loopquant}

\begin{algorithm}[h]
\caption{LoopQuant: Overall Quantization Pipeline}
\label{alg:loopquant}

\KwIn{
Pretrained LoopLM $\mathcal{M}$,
calibration dataset $\mathcal{D}$,
target activation bit-width budget $\overline b$
}

\KwOut{
Quantized LoopLM $\widehat{\mathcal{M}}$
}

\For{each recursively reused linear module $m$}{

\tcp{Stage 1: Cross-Loop Distribution Canonicalization (CLDC)}

Collect loop-wise calibration activations
$\{\mathbf X_m^{(t)}\}_{t=1}^{T}$\;

Initialize the shared canonical transformation\;

Optimize cross-loop reconstruction robustness,
anomaly synchronization, and group-aware reordering to obtain
$\mathbf P_m^\star
=
\mathbf D_m^\star
\mathbf R_m^\star
\mathbf \Pi_m^\star$\;

Apply the canonical transformation to obtain
$\widetilde{\mathbf W}_m$\;

Quantize the canonicalized weight with GPTQ:
$\widehat{\mathbf W}_m = Q_W(\widetilde{\mathbf W}_m)$\;
}

\BlankLine

\tcp{Stage 2: Interaction-Aware Loop-wise Precision Allocation (ILPA)}

Construct candidate activation precision configurations:
$b_t\in\{b_{\mathrm L},b_{\mathrm H}\}$\;

Evaluate trajectory-level task cost under recursive precision contexts\;

Estimate loop sensitivity $\phi_t$ and cross-loop interaction
$\widetilde{\Gamma}_{ij}$\;

Solve the precision assignment by exact enumeration for small $T$,
or greedy allocation otherwise, under budget $\overline b$\;

Obtain the final loop-wise precision configuration
$\{b_t^\star\}_{t=1}^{T}$\;

\BlankLine

\tcp{Stage 3: Loop-Conditioned Task Calibration (LCTC)}

Freeze quantized weights, CLDC transformations, and ILPA precision assignments\;

Introduce static loop-conditioned calibration parameters:
$\boldsymbol{\Gamma}_{m,t}
=
\{
\mathbf c_{m,t},
\mathbf g_{m,t},
\boldsymbol{\beta}_{m,t}
\}$\;

Optimize task-space calibration objective
$\mathcal L_{\mathrm{LCTC}}$\;

Remove all temporary optimization variables\;

Assemble the final deployable model:
$\widehat{\mathcal M}
=
\{
\widehat{\mathbf W}_m,
\mathbf P_m^\star,
\boldsymbol{\Gamma}_{m,t},
b_t^\star
\}$

\Return{$\widehat{\mathcal M}$}

\end{algorithm}

Algorithm~\ref{alg:loopquant} summarizes the overall workflow of LoopQuant.
The framework first canonicalizes recursive activation distributions through
CLDC, then allocates loop-wise activation precision with ILPA under a fixed
bit-width budget, and finally performs task-aligned calibration along the
low-bit recursive trajectory through LCTC.

\subsubsection{Cross-Loop Distribution Canonicalization (CLDC)}
\label{sec:code-cldc}

CLDC aims to construct a shared quantization-friendly representation across
recursive states while preserving the parameter-sharing structure of LoopLMs.
It first learns a shared canonical geometry from all loop-wise calibration
states, then reduces cross-loop distribution mismatch through robust
synchronization, and finally performs group-aware channel reordering to improve
group-wise quantization compatibility.

\begin{algorithm}[h]
\caption{Cross-Loop Distribution Canonicalization (CLDC)}
\label{alg:cldc}

\KwIn{
Shared linear weight $\mathbf W_m$,
loop-wise calibration activations
$\{\mathbf X_m^{(t)}\}_{t=1}^{T}$
}

\KwOut{
Canonicalized weight $\widetilde{\mathbf W}_m$,
shared transformation $\mathbf P_m^\star$
}

\tcp{Stage I: Shared canonical geometry learning}

Initialize the shared transformation:
$\mathbf P_m^{\mathrm{cont}}
=
\mathbf D_m\mathbf R_m$\;

Jointly optimize $\mathbf D_m$ and $\mathbf R_m$ using all loop calibration
states\;

Apply the shared transformation:
$\widetilde{\mathbf X}_m^{(t)}
=
\mathbf X_m^{(t)}
\mathbf D_m
\mathbf R_m$\;

Absorb the inverse transformation into the weight:
$\widetilde{\mathbf W}_m
=
\mathbf W_m
\mathbf D_m^{-1}
\mathbf R_m$\;

\BlankLine

\tcp{Stage II: Robust cross-loop synchronization}

Compute loop-wise quantization reconstruction error
$\mathcal E_m^{(t)}$\;

Optimize the smooth worst-loop objective
$\mathcal L_{\mathrm{rob}}$\;

Extract cross-loop anomaly profiles
$\mathbf p_m^{(t)}$\;

Optimize anomaly synchronization objective
$\mathcal L_{\mathrm{sync}}$\;

Obtain optimized continuous canonical geometry:
$(\mathbf D_m^\star,\mathbf R_m^\star)$\;

\BlankLine

\tcp{Stage III: Group-aware channel reordering}

Construct cross-loop channel signatures\;

Initialize channel permutation according to worst-case dynamic range\;

Optimize hard permutation:
$\mathbf\Pi_m^\star$\;

Construct final shared transformation:
$\mathbf P_m^\star
=
\mathbf D_m^\star
\mathbf R_m^\star
\mathbf\Pi_m^\star$\;

Apply final canonicalization to obtain
$\widetilde{\mathbf W}_m$\;

\Return{$\widetilde{\mathbf W}_m,\mathbf P_m^\star$}

\end{algorithm}

Algorithm~\ref{alg:cldc} summarizes the CLDC procedure.
CLDC first learns a shared canonical geometry across recursive states,
then aligns cross-loop activation distributions through robust synchronization,
and finally improves group-wise quantization compatibility with channel
reordering.
The resulting transformation is shared across all loops and remains fixed for
subsequent ILPA and LCTC stages.

\subsubsection{Interaction-Aware Loop-wise Precision Allocation (ILPA)}
\label{sec:code-ilpa}

ILPA determines loop-wise activation precision under a fixed bit-width budget.
It first evaluates the task impact of different precision configurations along
the recursive trajectory, then estimates loop sensitivity and cross-loop
interaction, and finally allocates activation precision through a
budget-aware optimization procedure.

\begin{algorithm}[h]
\caption{Interaction-Aware Loop-wise Precision Allocation (ILPA)}
\label{alg:ilpa}

\KwIn{
Quantized LoopLM,
calibration dataset $\mathcal D$,
low/high activation precision
$b_{\mathrm L},b_{\mathrm H}$,
target average bit-width $\overline b$
}

\KwOut{
Loop-wise precision configuration
$\{b_t^\star\}_{t=1}^{T}$
}

\tcp{Stage I: Trajectory-level precision evaluation}

Construct candidate precision configurations:
$\mathcal S\subseteq\{1,\ldots,T\}$\;

Evaluate recursive forward passes under different precision configurations\;

Compute task-level cost:
$v(\mathcal S)$\;

\BlankLine

\tcp{Stage II: Sensitivity and interaction estimation}

Estimate context-dependent loop sensitivity:
$\phi_t$\;

Estimate pairwise recursive interaction:
$\widetilde{\Gamma}_{ij}$\;

Construct the precision allocation objective:
$\mathcal J(\mathbf q)
=
\sum_t q_t\phi_t
+
\lambda\sum_{i<j}
q_iq_j\widetilde{\Gamma}_{ij}$\;

\BlankLine

\tcp{Stage III: Budget-aware precision allocation}

\eIf{$T$ is small}{
Enumerate feasible assignments satisfying the bit budget and select the
minimum-$\mathcal J$ configuration\;
}{
Initialize all loops with high activation precision:
$q_t=0$\;

\While{activation bit-width budget is not satisfied}{
Compute marginal precision reduction cost for each candidate loop\;

Select the loop with minimum marginal cost:
$t^\star=\arg\min_t\Delta\mathcal J_t$\;

Assign loop $t^\star$ to low precision:
$q_{t^\star}=1$\;
}
}

Obtain final precision assignment:
$\{b_t^\star\}_{t=1}^{T}$\;

\Return{$\{b_t^\star\}_{t=1}^{T}$}

\end{algorithm}

Algorithm~\ref{alg:ilpa} summarizes the ILPA procedure.
ILPA first evaluates precision configurations on the complete recursive
trajectory, then identifies loop-wise sensitivity and cross-loop interaction,
and finally allocates activation precision under the target bit-width budget.
The obtained precision configuration remains fixed during inference and is
used by the subsequent LCTC stage.

\subsubsection{Loop-Conditioned Task Calibration (LCTC)}
\label{sec:code-lctc}

LCTC performs the final task-level calibration after CLDC and ILPA.
It keeps the quantization structure and precision assignment fixed, and only
optimizes lightweight loop-conditioned calibration parameters along the
low-bit recursive trajectory.
The calibration objective jointly considers teacher alignment, language
modeling loss, clean-state task defects, and trajectory-dependent defects.

\begin{algorithm}[h]
\caption{Loop-Conditioned Task Calibration (LCTC)}
\label{alg:lctc}

\KwIn{
Quantized LoopLM,
CLDC transformations,
ILPA precision configuration,
teacher model $\mathcal M$,
calibration dataset $\mathcal D$
}

\KwOut{
Final deployable LoopQuant model
$\widehat{\mathcal M}$
}

\tcp{Stage I: Loop-conditioned calibration initialization}

Freeze quantized weights, CLDC transformations, and ILPA precision assignment\;

Initialize static calibration parameters:
$\boldsymbol{\Gamma}_{m,t}
=
\{
\mathbf c_{m,t},
\mathbf g_{m,t},
\boldsymbol{\beta}_{m,t}
\}$\;

\BlankLine

\tcp{Stage II: Task-space defect calibration}

Collect low-bit recursive trajectories\;

Compute clean-state task defects:
$\mathbf a_t$\;

Compute trajectory-dependent defects:
$\mathbf s_t(\alpha)$\;

Evaluate Fisher-weighted task losses:
$\mathcal L_{\mathrm{anchor}}$ and
$\mathcal L_{\mathrm{traj}}$\;

\BlankLine

\tcp{Stage III: Task-aligned optimization}

Optimize loop-conditioned calibration parameters:

$\boldsymbol{\Gamma}_{m,t}
\leftarrow
\arg\min
\mathcal L_{\mathrm{LCTC}}$\;

Keep quantized weights, CLDC transformations, ILPA precision assignments,
and backbone parameters frozen\;

Remove temporary optimization variables\;

Assemble final deployable model:
$\widehat{\mathcal M}
=
\{
\widehat{\mathbf W}_m,
\mathbf P_m^\star,
\boldsymbol{\Gamma}_{m,t},
b_t^\star
\}$\;

\Return{$\widehat{\mathcal M}$}

\end{algorithm}

Algorithm~\ref{alg:lctc} summarizes the final calibration procedure of
LoopQuant.
LCTC does not modify the quantization structure or precision allocation.
Instead, it optimizes lightweight loop-conditioned parameters to correct
task-space quantization defects and trajectory-dependent error amplification.
After calibration, only static calibration metadata are retained in addition to
the quantized model; temporary optimization states are discarded.

\subsection{Detailed Derivations and Implementation Details}
\label{sec:proof}

\subsubsection{Orthogonal Canonical Geometry Parameterization}
\label{app:cldc_geometry}

The goal of CLDC is to learn a shared transformation that improves quantization compatibility across all recurrent states while preserving the original computation of LoopLMs. 
For a recursively reused linear module with shared weight $\mathbf{W}_m\in\mathbb{R}^{d_o\times d_i}$, let $\mathbf{X}_m^{(t)}$ denote the input activation collected from the $t$-th loop. 
A naive approach would optimize an independent transformation $\mathbf{P}_m^{(t)}$ for each loop. 
Although this formulation can better fit individual activation distributions, it violates the parameter-sharing property of LoopLMs and introduces additional deployment overhead. 
Therefore, CLDC seeks a single canonical transformation shared across all recurrent states:

\begin{equation}
\mathbf{P}_m^{(1)}
=
\mathbf{P}_m^{(2)}
=
\cdots
=
\mathbf{P}_m^{(T)}
\triangleq
\mathbf{P}_m .
\end{equation}

We decompose the continuous canonical transformation into a channel-wise scaling component and an orthogonal rotation component:

\begin{equation}
\mathbf{P}_m^{\mathrm{cont}}
=
\mathbf{D}_m\mathbf{R}_m,
\qquad
\mathbf{D}_m=\operatorname{Diag}(\mathbf{s}_m),
\qquad
\mathbf{s}_m>0,
\end{equation}

where $\mathbf{D}_m$ adjusts the channel-wise dynamic range and $\mathbf{R}_m$ redistributes cross-channel geometry while preserving the Euclidean structure. 
The orthogonality constraint is imposed as

\begin{equation}
\mathbf{R}_m\mathbf{R}_m^{\top}=\mathbf{I}.
\end{equation}

Directly optimizing a dense orthogonal matrix introduces a large number of constrained parameters and makes optimization inefficient. 
To obtain an expressive yet tractable parameterization, we represent the rotation matrix as

\begin{equation}
\mathbf{R}_m
=
\mathbf{R}_m^{(0)}
\mathbf{U}_m ,
\end{equation}

where $\mathbf{R}_m^{(0)}$ is a fixed structured orthogonal transformation and $\mathbf{U}_m$ is a learnable orthogonal correction. 
Instead of optimizing a full dense matrix, we adopt a block-diagonal parameterization:

\begin{equation}
\mathbf{U}_m
=
\operatorname{BlkDiag}
(
\mathbf{U}_{m,1},
\ldots,
\mathbf{U}_{m,B}
).
\end{equation}

For the $b$-th block, we introduce an unconstrained matrix $\mathbf{Z}_{m,b}$ and construct its corresponding skew-symmetric matrix:

\begin{equation}
\mathbf{A}_{m,b}
=
\mathbf{Z}_{m,b}
-
\mathbf{Z}_{m,b}^{\top}.
\end{equation}

The orthogonal correction is then obtained through the Cayley transform:

\begin{equation}
\mathbf{U}_{m,b}
=
\left(
\mathbf{I}
-
\frac{1}{2}\mathbf{A}_{m,b}
\right)^{-1}
\left(
\mathbf{I}
+
\frac{1}{2}\mathbf{A}_{m,b}
\right).
\end{equation}

Since $\mathbf{A}_{m,b}^{\top}=-\mathbf{A}_{m,b}$, the Cayley transform guarantees

\begin{equation}
\mathbf{U}_{m,b}
\mathbf{U}_{m,b}^{\top}
=
\mathbf{I},
\end{equation}

and consequently

\begin{equation}
\mathbf{R}_m
\mathbf{R}_m^{\top}
=
\mathbf{R}_m^{(0)}
\mathbf{U}_m
\mathbf{U}_m^{\top}
(\mathbf{R}_m^{(0)})^{\top}
=
\mathbf{I}.
\end{equation}

Therefore, the learned rotation remains strictly orthogonal throughout optimization without requiring explicit projection onto the orthogonal manifold.

We initialize all learnable correction matrices as $\mathbf{Z}_{m,b}=\mathbf{0}$. 
At initialization, this gives

\begin{equation}
\mathbf{U}_{m,b}=\mathbf{I},
\qquad
\mathbf{R}_m=\mathbf{R}_m^{(0)},
\end{equation}

meaning that optimization starts from the original structured rotation space and only learns data-supported orthogonal corrections from cross-loop calibration states.

For every loop, CLDC applies the same transformation to activations and compensates the inverse transformation in weights:

\begin{equation}
\widetilde{\mathbf{X}}_m^{(t)}
=
\mathbf{X}_m^{(t)}
\mathbf{D}_m
\mathbf{R}_m,
\qquad
\widetilde{\mathbf{W}}_m
=
\mathbf{W}_m
\mathbf{D}_m^{-1}
\mathbf{R}_m .
\end{equation}

The transformed computation satisfies

\begin{equation}
\begin{aligned}
\widetilde{\mathbf{X}}_m^{(t)}
\widetilde{\mathbf{W}}_m^{\top}
&=
\mathbf{X}_m^{(t)}
\mathbf{D}_m
\mathbf{R}_m
\mathbf{R}_m^{\top}
\mathbf{D}_m^{-1}
\mathbf{W}_m^{\top}
\\
&=
\mathbf{X}_m^{(t)}
\mathbf{W}_m^{\top},
\qquad
\forall t ,
\end{aligned}
\end{equation}

which proves that CLDC preserves the original full-precision function while modifying only the quantization geometry.

\subsubsection{Robust Cross-Loop Synchronization Objective}
\label{app:cldc_sync}

After obtaining the shared canonical geometry, CLDC further optimizes the transformation by jointly considering the quantization behavior across all recurrent states. 
For the $t$-th loop, we measure the quantization distortion of the transformed representation using the normalized reconstruction error:

\begin{equation}
\mathcal{E}_m^{(t)}
=
\frac{
\left\|
Q_A(\widetilde{\mathbf{X}}_m^{(t)})
Q_W(\widetilde{\mathbf{W}}_m)^{\top}
-
\mathbf{X}_m^{(t)}
\mathbf{W}_m^{\top}
\right\|_F^2
}
{
\left\|
\mathbf{X}_m^{(t)}
\mathbf{W}_m^{\top}
\right\|_F^2
},
\end{equation}

where $Q_A(\cdot)$ and $Q_W(\cdot)$ denote the activation and weight quantizers under the target quantization configuration, respectively. 
This reconstruction error serves as a local approximation of the loop-dependent quantization perturbation $\boldsymbol{\epsilon}^{\mathrm{loc}}_t$ introduced in Sec.~\ref{sec:formulate}.

A straightforward objective is to minimize the average reconstruction error:

\begin{equation}
\mathcal{L}_{\mathrm{avg}}
=
\frac{1}{T}
\sum_{t=1}^{T}
\mathcal{E}_m^{(t)} .
\end{equation}

However, this objective treats all recurrent states equally and may allow a small number of difficult loops to be dominated by easily quantized states. 
Such behavior is undesirable in LoopLMs because severe distortion at an early recurrent position can propagate through subsequent loops and accumulate along the recursive trajectory.

To explicitly control the worst-case quantization risk across loops, we introduce a smooth worst-loop objective:

\begin{equation}
\mathcal{L}_{\mathrm{rob}}
=
\tau
\log
\left(
\frac{1}{T}
\sum_{t=1}^{T}
\exp
\left(
\frac{\mathcal{E}_m^{(t)}}{\tau}
\right)
\right),
\end{equation}

where $\tau$ controls the smoothness of the approximation. 
When $\tau$ approaches zero, this objective converges to the maximum loop-wise error:

\begin{equation}
\lim_{\tau\rightarrow0}
\mathcal{L}_{\mathrm{rob}}
=
\max_t
\mathcal{E}_m^{(t)} .
\end{equation}

Therefore, $\mathcal{L}_{\mathrm{rob}}$ encourages the shared canonical geometry to reduce the largest quantization risk among all recurrent states rather than optimizing only the average behavior.

Nevertheless, reducing reconstruction error does not guarantee that different loops exhibit consistent activation outlier structures. 
After transformation, different recurrent states may still produce shifted anomaly channels, which can degrade group-wise quantization because channels with different dynamic ranges may share identical quantization parameters.

To characterize the cross-loop anomaly dynamics, we define the high-percentile activation magnitude of channel $c$ at loop $t$ as

\begin{equation}
a_{m,c}^{(t)}
=
\operatorname{Quantile}_{q}
\left(
\left|
\widetilde{\mathbf{X}}_{m,:,c}^{(t)}
\right|
\right),
\end{equation}

where $q$ denotes a high percentile threshold. 
The anomaly statistic is further normalized across channels:

\begin{equation}
p_{m,c}^{(t)}
=
\frac{
a_{m,c}^{(t)}
}
{
\sum_{k=1}^{d_i}
a_{m,k}^{(t)}
}.
\end{equation}

The resulting vector $\mathbf{p}_m^{(t)}$ represents the relative distribution of activation anomalies across channels for loop $t$. 
Instead of selecting a specific loop as a reference state, CLDC assumes that all recurrent states should share a consistent anomaly structure. 
Therefore, we define the cross-loop anomaly barycenter as

\begin{equation}
\overline{\mathbf{p}}_m
=
\arg\min_{\mathbf{p}}
\frac{1}{T}
\sum_{t=1}^{T}
\|
\mathbf{p}_m^{(t)}
-
\mathbf{p}
\|_2^2 .
\end{equation}

The above objective has a closed-form solution:

\begin{equation}
\overline{\mathbf{p}}_m
=
\frac{1}{T}
\sum_{t=1}^{T}
\mathbf{p}_m^{(t)} .
\end{equation}

Based on this shared anomaly center, we define the cross-loop synchronization loss:

\begin{equation}
\mathcal{L}_{\mathrm{sync}}
=
\frac{1}{T}
\sum_{t=1}^{T}
\|
\mathbf{p}_m^{(t)}
-
\overline{\mathbf{p}}_m
\|_2^2 .
\end{equation}

Unlike choosing one loop as an artificial anchor, this formulation jointly reduces the distribution discrepancy among all recurrent states and encourages different loops to share similar anomaly structures in the learned canonical space.

Finally, because channel scaling can arbitrarily enlarge activation scales while reducing activation quantization error, excessive scaling may transfer quantization difficulty from activations to weights. 
We therefore introduce a scale regularization term:

\begin{equation}
\mathcal{L}_{\mathrm{reg}}
=
\frac{1}{d_i}
\|
\mathbf{s}_m-\mathbf{1}
\|_2^2 .
\end{equation}

Combining reconstruction robustness, anomaly synchronization, and scale regularization, the final optimization objective for the shared canonical geometry is:

\begin{equation}
(\mathbf{D}_m^\star,\mathbf{R}_m^\star)
=
\arg\min
\left[
\mathcal{L}_{\mathrm{rob}}
+
\lambda_{\mathrm{sync}}
\mathcal{L}_{\mathrm{sync}}
+
\lambda_{\mathrm{reg}}
\mathcal{L}_{\mathrm{reg}}
\right].
\end{equation}

The optimized transformation therefore represents a consensus quantization geometry determined by the complete recursive trajectory, rather than a local optimum for any individual loop.

\subsubsection{Group-Aware Channel Reordering}
\label{app:cldc_reorder}

Although the continuous canonical geometry improves the overall quantization compatibility across recurrent states, group-wise quantization introduces an additional constraint: channels within the same quantization group must share identical quantization parameters. 
Therefore, even after canonicalization, channels with inconsistent dynamic ranges across loops may still be assigned to the same group and introduce intra-group quantization mismatch.

To address this issue, CLDC performs a discrete channel reordering step after optimizing the continuous transformation. 
Unlike continuous permutation relaxation, which introduces additional optimization variables and deployment overhead, we formulate reordering as a hard group assignment problem that directly optimizes the final quantization layout.

After fixing the optimized scaling and rotation matrices $\mathbf{D}_m^\star$ and $\mathbf{R}_m^\star$, we first characterize each channel using its complete cross-loop dynamic behavior. 
For channel $c$ at loop $t$, we define the logarithmic range signature as

\begin{equation}
z_{m,c}^{(t)}
=
\log
\left(
a_{m,c}^{(t)}
+
\varepsilon
\right),
\end{equation}

where $a_{m,c}^{(t)}$ is the anomaly statistic defined in Appendix~\ref{app:cldc_sync} and $\varepsilon$ is a small constant for numerical stability. 
The complete cross-loop signature of channel $c$ is represented as

\begin{equation}
\mathbf{z}_{m,c}
=
\left[
z_{m,c}^{(1)},
z_{m,c}^{(2)},
\ldots,
z_{m,c}^{(T)}
\right].
\end{equation}

A channel permutation matrix $\mathbf{\Pi}_m$ divides input channels into $K$ consecutive quantization groups:

\begin{equation}
\{
\mathcal{G}_{m,1}(\mathbf{\Pi}_m),
\ldots,
\mathcal{G}_{m,K}(\mathbf{\Pi}_m)
\}.
\end{equation}

For a given loop $t$, we measure the quantization incompatibility inside each group by the variance of channel signatures:

\begin{equation}
\mathcal{C}_m^{(t)}(\mathbf{\Pi}_m)
=
\frac{1}{K}
\sum_{k=1}^{K}
\operatorname{Var}
\left(
\left\{
z_{m,c}^{(t)}
:
c\in
\mathcal{G}_{m,k}(\mathbf{\Pi}_m)
\right\}
\right).
\end{equation}

A lower value indicates that channels within each group have more consistent dynamic ranges and are therefore more suitable for sharing quantization parameters.

Similar to the robust optimization in CLDC, optimizing only the average group variance may hide difficult recurrent states. 
Therefore, we define a cross-loop robust group objective:

\begin{equation}
\mathcal{L}_{\mathrm{grp}}
(\mathbf{\Pi}_m)
=
\tau_g
\log
\left(
\frac{1}{T}
\sum_{t=1}^{T}
\exp
\left(
\frac{
\mathcal{C}_m^{(t)}(\mathbf{\Pi}_m)
}
{\tau_g}
\right)
\right).
\end{equation}

The optimal channel ordering is obtained by solving the discrete optimization problem:

\begin{equation}
\mathbf{\Pi}_m^\star
=
\arg\min_{\mathbf{\Pi}\in\mathcal{P}_{d_i}}
\mathcal{L}_{\mathrm{grp}}
(\mathbf{\Pi}),
\end{equation}

where $\mathcal{P}_{d_i}$ denotes the set of all $d_i$-dimensional permutation matrices.

Directly searching the permutation space is computationally infeasible due to its combinatorial size. 
Therefore, we adopt a two-stage optimization strategy. 
First, we construct an initialization by sorting channels according to their worst-case dynamic range:

\begin{equation}
\rho_{m,c}
=
\max_t
a_{m,c}^{(t)} ,
\end{equation}

and assign channels according to the ordering of $\rho_{m,c}$ to obtain an initial permutation $\mathbf{\Pi}_m^{(0)}$.

Starting from this initialization, we further perform local pairwise-swap refinement. 
At each step, two channels from different quantization groups are exchanged, and the update is accepted only when it decreases $\mathcal{L}_{\mathrm{grp}}(\mathbf{\Pi}_m)$. 
This procedure preserves a hard permutation structure and avoids introducing additional continuous parameters during deployment.

Finally, combining the optimized continuous canonical geometry and discrete
group-aware ordering, CLDC obtains the final shared transformation:

\begin{equation}
\mathbf{P}_m^\star
=
\mathbf{D}_m^\star
\mathbf{R}_m^\star
\mathbf{\Pi}_m^\star .
\end{equation}

The corresponding transformed activation and weight are defined as

\begin{equation}
\widetilde{\mathbf X}^{(t)}_m
=
\mathbf X^{(t)}_m
\mathbf P_m^\star,
\qquad
\widetilde{\mathbf W}_m
=
\mathbf W_m
(\mathbf P_m^\star)^{-\top}.
\end{equation}

Since $\mathbf{\Pi}_m^\star$ is an orthogonal permutation matrix, the final
transformation preserves the original full-precision computation:

\begin{equation}
\widetilde{\mathbf X}^{(t)}_m
\widetilde{\mathbf W}_m^{\top}
=
\mathbf X^{(t)}_m
\mathbf W_m^{\top},
\quad \forall t .
\end{equation}

The resulting permutation is shared across all recurrent states, ensuring
that the quantization layout remains consistent with the parameter-sharing
design of LoopLMs.

\subsubsection{Shapley-Style Loop Sensitivity Estimation}
\label{app:ilpa_sensitivity}

The goal of ILPA is to estimate the task-level sensitivity of each loop to activation precision reduction. 
Unlike layer-wise quantization sensitivity, the importance of a loop in LoopLMs depends on the precision states of other loops because quantization perturbations propagate through the recursive trajectory. 
Therefore, we evaluate each loop under different recursive precision contexts.

Let $\mathcal{T}=\{1,\ldots,T\}$ denote all loops. 
For a precision configuration $\mathcal{S}\subseteq\mathcal{T}$, the loops in $\mathcal{S}$ are assigned low activation precision $b_{\mathrm L}$, while the remaining loops use high activation precision $b_{\mathrm H}$. 
The corresponding precision assignment is

\begin{equation}
b_t(\mathcal{S})
=
\begin{cases}
b_{\mathrm L}, & t\in\mathcal{S},\\
b_{\mathrm H}, & t\notin\mathcal{S}.
\end{cases}
\end{equation}

Given a precision configuration $\mathcal{S}$, the model executes a complete recursive forward pass with the selected activation precision assignment. 
The task-level cost is defined as the calibration-set token-level negative log-likelihood:

\begin{equation}
v(\mathcal{S})
=
\mathbb{E}_{\mathbf{x}\sim\mathcal{D}_{\mathrm{cal}}}
\left[
-\frac{1}{|\mathbf{x}|}
\sum_n
\log p_{\mathcal{S}}(x_{n+1}|x_{\leq n})
\right].
\end{equation}

This trajectory-level objective differs from local reconstruction metrics because it directly measures the influence of precision decisions on the final model behavior.

To estimate the marginal contribution of each loop, we consider a permutation
$\pi\in\mathfrak{S}_T$, where $\mathfrak{S}_T$ denotes the set of all permutations over $T$ loops. 
For a given permutation, let $\mathcal{S}_{t}^{(\pi)}$ denote the set of loops appearing before loop $t$ and already assigned low precision. 
The marginal degradation caused by additionally reducing the precision of loop $t$ is defined as

\begin{equation}
\Delta_t^{(\pi)}
=
v(\mathcal{S}_{t}^{(\pi)}\cup\{t\})
-
v(\mathcal{S}_{t}^{(\pi)}).
\end{equation}

The above quantity measures the additional task degradation introduced by loop $t$ under a specific recursive precision context. 
Since different permutations correspond to different combinations of preceding low-precision loops, averaging over all permutations provides a context-independent estimation of loop importance.

Following the Shapley value formulation, the loop sensitivity is computed as

\begin{equation}
\phi_t
=
\frac{1}{T!}
\sum_{\pi\in\mathfrak{S}_T}
\Delta_t^{(\pi)} .
\end{equation}

A larger $\phi_t$ indicates that reducing the precision of loop $t$ consistently introduces larger task degradation across different recursive contexts. 
Therefore, $\phi_t$ provides a discrete task-level estimate of the first-order sensitivity induced by reducing the precision of loop $t$.

In practice, enumerating all $T!$ permutations is unnecessary. 
Instead, ILPA samples a subset of permutations and estimates the expectation using Monte Carlo approximation:

\begin{equation}
\phi_t
\approx
\frac{1}{M}
\sum_{m=1}^{M}
\Delta_t^{(\pi_m)},
\end{equation}

where $\{\pi_m\}_{m=1}^{M}$ are randomly sampled loop permutations. 
The sampled estimation provides a scalable approximation while preserving the context-aware sensitivity property required for loop-wise precision allocation.

\subsubsection{Recursive Interaction Modeling}
\label{app:ilpa_interaction}

The sensitivity term $\phi_t$ in Appendix~\ref{app:ilpa_sensitivity} measures
the individual contribution of each loop when its activation precision is
reduced. 
However, the recursive structure of LoopLMs introduces additional coupling
effects: two loops that are individually insensitive may still cause
significant degradation when simultaneously quantized at low precision
because their perturbations interact along the recursive trajectory. 
Therefore, ILPA explicitly models the pairwise interaction among loops to
capture non-additive task effects under discrete precision perturbations.

For two loops $i$ and $j$, we first define the joint marginal degradation
under a recursive precision context. 
Given a permutation $\pi\in\mathfrak{S}_T$, let
$\mathcal{S}^{(\pi)}$ denote the set of loops assigned low precision before
evaluating the joint contribution of loops $i$ and $j$. 
The joint marginal degradation is defined as

\begin{equation}
\Delta_{ij}^{(\pi)}
=
v(\mathcal{S}^{(\pi)}\cup\{i,j\})
-
v(\mathcal{S}^{(\pi)}).
\end{equation}

The individual marginal degradations of loops $i$ and $j$ under the same
context are

\begin{equation}
\Delta_i^{(\pi)}
=
v(\mathcal{S}^{(\pi)}\cup\{i\})
-
v(\mathcal{S}^{(\pi)}),
\end{equation}

\begin{equation}
\Delta_j^{(\pi)}
=
v(\mathcal{S}^{(\pi)}\cup\{j\})
-
v(\mathcal{S}^{(\pi)}).
\end{equation}

If two loops contribute independently, their joint degradation should
approximately equal the sum of individual degradations. 
Therefore, we define the additional recursive interaction as the
non-additive component:

\begin{equation}
\Gamma_{ij}
=
\frac{1}{T!}
\sum_{\pi\in\mathfrak{S}_T}
\left(
\Delta_{ij}^{(\pi)}
-
\Delta_i^{(\pi)}
-
\Delta_j^{(\pi)}
\right).
\end{equation}

A positive $\Gamma_{ij}$ indicates that jointly reducing the precision of
loops $i$ and $j$ introduces stronger-than-additive degradation, implying
that their quantization perturbations interact along the recursive
trajectory. 
In contrast, a value close to zero indicates that the two loops behave
approximately independently under precision reduction.

The interaction term $\Gamma_{ij}$ captures the discrete task-level
interaction effect induced by jointly reducing the precision of loops $i$
and $j$. 
While the second-order formulation in Sec.~\ref{sec:formulate} motivates
modeling interactions among perturbations injected at different loops,
$\Gamma_{ij}$ measures the corresponding non-additive task degradation under
discrete precision configurations.

The above formulation connects the trajectory-level objective $v(\mathcal S)$
with ILPA's discrete allocation objective. 
Starting from the second-order perturbation analysis in
Sec.~\ref{sec:formulate}, precision reduction can be viewed as a discrete
perturbation variable. 
The first-order contribution is captured by the marginal task degradation
$\phi_t$, while the additional interaction among multiple precision
reductions is captured by $\Gamma_{ij}$. 
Therefore, ILPA formulates $\mathcal{J}(\mathbf q)$ as a discrete
second-order surrogate of the original trajectory-level objective, enabling
efficient precision allocation without exhaustively evaluating all possible
precision configurations.

Since different loop pairs may exhibit different degradation scales across
recursive precision contexts, directly comparing $\Gamma_{ij}$ may introduce
magnitude bias. 
We therefore normalize the interaction strength using the empirical variance
of marginal degradation:

\begin{equation}
\widetilde{\Gamma}_{ij}
=
\frac{\Gamma_{ij}}
{\sqrt{\sigma_i^2\sigma_j^2+\epsilon}},
\end{equation}

where $\sigma_i^2$ and $\sigma_j^2$ are estimated as

\begin{equation}
\sigma_i^2
=
\frac{1}{|\Pi|}
\sum_{\pi\in\Pi}
\left(
\Delta_i^{(\pi)}
-
\phi_i
\right)^2,
\end{equation}

\begin{equation}
\sigma_j^2
=
\frac{1}{|\Pi|}
\sum_{\pi\in\Pi}
\left(
\Delta_j^{(\pi)}
-
\phi_j
\right)^2,
\end{equation}

where $\Pi$ denotes the sampled permutation set used for context estimation.

The normalization removes the influence of different marginal degradation
scales and makes $\widetilde{\Gamma}_{ij}$ comparable across different loop
pairs. 
The resulting normalized interaction matrix $\widetilde{\Gamma}$ is then
used in ILPA's precision allocation objective to model recursive coupling
among low-precision loops.

\subsubsection{Exact and Greedy Precision Allocation}
\label{app:ilpa_solver}

After estimating loop sensitivity $\phi_t$ and normalized interaction $\widetilde{\Gamma}_{ij}$, ILPA determines the final activation precision configuration under a fixed bit-width budget. 
The precision allocation objective is

\begin{equation}
\mathcal{J}(\mathbf q)
=
\sum_{t=1}^{T}
q_t\phi_t
+
\lambda
\sum_{i<j}
q_iq_j\widetilde{\Gamma}_{ij},
\end{equation}

where $q_t=1$ indicates low precision and $q_t=0$ high precision. 
The first term measures individual loop sensitivity, while the second captures additional degradation from interactions among low-precision loops.

The assignment is constrained by the target average activation bit-width:

\begin{equation}
\sum_{t=1}^{T}
\left[
b_{\mathrm H}
-
(b_{\mathrm H}-b_{\mathrm L})q_t
\right]
=
T\overline b .
\end{equation}

For small $T$, we enumerate all feasible binary assignments satisfying the budget and select the configuration with the minimum $\mathcal{J}(\mathbf q)$, yielding the exact optimum under the estimated objective. 
As $T$ increases, exhaustive search over up to $2^T$ assignments becomes expensive. 
We therefore use a greedy reduction strategy that progressively converts high-precision loops into low-precision loops.

Starting from

\begin{equation}
q_t=0,
\qquad
\forall t,
\end{equation}

we evaluate the marginal objective increase from reducing loop $t$ to low precision:

\begin{equation}
\Delta\mathcal{J}_t
=
\phi_t
+
\lambda
\sum_{j:q_j=1}
\widetilde{\Gamma}_{tj}.
\end{equation}

The next loop is selected as

\begin{equation}
t^\star
=
\arg\min_{t:q_t=0}
\Delta\mathcal{J}_t ,
\end{equation}

after which $q_{t^\star}=1$. 
The process repeats until the bit-width constraint is satisfied, prioritizing loops with smaller task sensitivity and weaker interactions for low-bit execution.

The greedy solver requires at most $T$ selection steps and, with cached sensitivity and interaction values, has overall complexity $\mathcal{O}(T^2)$. 
It avoids exponential search while preserving the task-aware allocation of ILPA.

\subsubsection{Loop-Conditioned Calibration Parameterization}
\label{app:lctc_param}

After CLDC and ILPA, the shared quantization transformation and loop-wise precision assignment are fixed. 
LCTC therefore adjusts only lightweight loop-conditioned calibration parameters while leaving quantized weights and backbone parameters unchanged. 
For module $m$ at loop $t$, we define

\begin{equation}
\boldsymbol{\Gamma}_{m,t}
=
\left\{
\mathbf c_{m,t},
\mathbf g_{m,t},
\boldsymbol{\beta}_{m,t}
\right\},
\end{equation}

where $\mathbf c_{m,t}$ controls group-wise activation clipping, while $\mathbf g_{m,t}$ and $\boldsymbol{\beta}_{m,t}$ provide group-wise gain and bias correction, respectively. 
Given the fixed quantized weight $\widehat{\mathbf W}_m$ and the activation precision $b_t$ selected by ILPA, the calibrated mapping is

\begin{equation}
\widehat{\mathbf y}_{m,t}
=
\mathbf g_{m,t}
\odot
\left[
\widehat{\mathbf W}_m
Q_{b_t}
\left(
\mathbf x_{m,t};
\mathbf c_{m,t}
\right)
+
\mathbf b_m
\right]
+
\boldsymbol{\beta}_{m,t}.
\end{equation}

The calibration parameters depend only on the physical module and loop index, not on individual tokens or inference histories. 
They are therefore stored as static lookup parameters and introduce no token-wise state cache or online optimization during inference. 
This parameterization changes only the loop-conditioned quantized response and preserves the quantization structure determined by CLDC and ILPA.

\subsubsection{Task-Space Defect Transport}
\label{app:lctc_transport}

The objective of LCTC is to calibrate quantization responses according to
their actual impact on the final task output rather than minimizing
intermediate hidden-state errors.
To achieve this goal, we explicitly transport loop-wise quantization defects
from hidden-state space into task space.

For the $t$-th loop, we denote the full-precision recursive computation as
$\mathcal{F}_{\theta,t}$ and the loop-conditioned quantized computation as
$\widehat{\mathcal F}_{t,\boldsymbol{\Gamma}}$.
Let $\mathcal{S}_{t+1}(\cdot)$ denote the suffix operator that maps the output
state of loop $t$ through the remaining full-precision recursive computation
and the final language modeling head:

\begin{equation}
\mathcal{S}_{t+1}(\mathbf h_t)
=
\operatorname{LMHead}
\left(
\mathcal{F}_{\theta,T}
\circ
\cdots
\circ
\mathcal{F}_{\theta,t+1}
(\mathbf h_t)
\right).
\end{equation}

Similarly, the full suffix mapping starting from the input state of loop $t$
is defined as

\begin{equation}
\mathcal{S}_{t}(\mathbf h_{t-1})
=
\mathcal{S}_{t+1}
\left(
\mathcal{F}_{\theta,t}
(\mathbf h_{t-1})
\right).
\end{equation}

Given an input state $\mathbf h$, the task-space quantization defect
introduced by loop $t$ is defined as

\begin{equation}
\mathbf d_t
(
\mathbf h;\boldsymbol{\Gamma}
)
=
\mathcal{S}_{t+1}
\left(
\widehat{\mathcal F}_{t,\boldsymbol{\Gamma}}
(\mathbf h)
\right)
-
\mathcal{S}_{t}
(\mathbf h).
\end{equation}

This formulation replaces only the current loop with the quantized operator
while keeping subsequent computation in full precision.
It is designed to isolate the task-space contribution of the current loop's
quantization error rather than reproduce the final deployed trajectory.
The complete low-bit recursive behavior is jointly optimized through the
teacher alignment and language modeling objectives in
Eq.~\eqref{eq:lctc_loss}.

Let $\mathbf H^{(t-1)}$ denote the hidden state of the full-precision
trajectory before loop $t$.
The clean-state task defect is defined as

\begin{equation}
\mathbf a_t
=
\mathbf d_t
(
\mathbf H^{(t-1)};
\boldsymbol{\Gamma}
).
\end{equation}

Although $\mathbf a_t$ measures the direct task impact of loop $t$ under its
clean input state, the actual quantized model does not arrive at loop $t$
through the full-precision trajectory.
Instead, previous quantization errors accumulate and introduce a trajectory
drift:

\begin{equation}
\boldsymbol{\delta}_{t-1}
=
\widehat{\mathbf H}^{(t-1)}
-
\mathbf H^{(t-1)} .
\end{equation}

The inherited drift changes the input distribution encountered by the
current loop and therefore modifies its quantization response.
To capture this effect without storing historical token-level states, LCTC
evaluates the directional response of the task-space defect along the
inherited drift direction.

Specifically, we define the task-space secant as

\begin{equation}
\mathbf s_t(\alpha)
=
\frac{
\mathbf d_t
(
\mathbf H^{(t-1)}
+
\alpha\boldsymbol{\delta}_{t-1};
\boldsymbol{\Gamma}
)
-
\mathbf d_t
(
\mathbf H^{(t-1)};
\boldsymbol{\Gamma}
)
}
{\alpha}.
\end{equation}

The secant formulation approximates the first-order sensitivity of the
task-space defect to the inherited trajectory deviation.
By applying the first-order Taylor expansion of $\mathbf d_t$ around
$\mathbf H^{(t-1)}$, we obtain

\begin{equation}
\mathbf d_t
(
\mathbf H^{(t-1)}
+
\alpha\boldsymbol{\delta}_{t-1};
\boldsymbol{\Gamma}
)
=
\mathbf d_t
(
\mathbf H^{(t-1)};
\boldsymbol{\Gamma}
)
+
\alpha
\mathbf J_{\mathbf d_t}
(
\mathbf H^{(t-1)}
)
\boldsymbol{\delta}_{t-1}
+
\mathcal{O}
(
\alpha^2
),
\end{equation}

where $\mathbf J_{\mathbf d_t}$ denotes the Jacobian of the task-space
defect with respect to the recurrent state.

Substituting the above expansion into the secant definition gives

\begin{equation}
\mathbf s_t(\alpha)
=
\mathbf J_{\mathbf d_t}
(
\mathbf H^{(t-1)}
)
\boldsymbol{\delta}_{t-1}
+
\mathcal{O}
(
\alpha
).
\end{equation}

Therefore, when $\alpha\rightarrow0$,

\begin{equation}
\lim_{\alpha\rightarrow0}
\mathbf s_t(\alpha)
=
\mathbf J_{\mathbf d_t}
(
\mathbf H^{(t-1)}
)
\boldsymbol{\delta}_{t-1}.
\end{equation}

This result shows that the secant term captures how inherited trajectory drift
is amplified by the current loop's quantization response in task space.
Therefore, $\mathbf s_t(\alpha)$ provides a first-order approximation of the
trajectory-dependent error
$\boldsymbol{\epsilon}^{\mathrm{traj}}_t$
introduced in Sec.~\ref{sec:formulate}.

For the first loop, no previous quantization perturbation exists, resulting in

\begin{equation}
\boldsymbol{\delta}_0=\mathbf 0,
\end{equation}

and consequently the trajectory-dependent term naturally becomes zero.
This property enables LCTC to distinguish direct loop-wise quantization
defects from errors accumulated through recursive computation.

\subsubsection{Fisher-Weighted Trajectory Calibration}
\label{app:lctc_fisher}

The goal of LCTC is to optimize lightweight calibration parameters according
to their impact on the final prediction distribution.
However, logit perturbations with identical Euclidean magnitude may lead to
different prediction changes because the output probability distribution is
highly non-uniform.
Therefore, LCTC adopts a Fisher-weighted metric to measure task-space defects
in a distribution-aware manner.

Given the teacher logits $\mathbf z\in\mathbb R^V$, where $V$ denotes the
vocabulary size, the corresponding probability distribution is

\begin{equation}
\mathbf p
=
\operatorname{softmax}(\mathbf z).
\end{equation}

For categorical distributions, the Fisher information matrix with respect to
the logits is defined as

\begin{equation}
\mathbf F(\mathbf p)
=
\operatorname{Diag}(\mathbf p)
-
\mathbf p\mathbf p^{\top}.
\end{equation}

For an arbitrary logit perturbation $\mathbf v$, the Fisher energy is defined
as

\begin{equation}
\mathcal E_{\mathrm F}(\mathbf v)
=
\mathbf v^{\top}
\mathbf F(\mathbf p)
\mathbf v .
\end{equation}

This metric measures the perturbation magnitude under the local geometry of
the prediction distribution rather than the Euclidean space.
To show its connection with task-level prediction change, we consider the KL
divergence between the teacher distribution and the perturbed distribution:

\begin{equation}
D_{\mathrm{KL}}
\left(
\mathbf p
\|
\operatorname{softmax}
(\mathbf z+\mathbf v)
\right).
\end{equation}

Using the second-order Taylor expansion around
$\mathbf v=\mathbf 0$, the first-order term vanishes because the teacher
distribution is the local optimum of the KL divergence.
Therefore,

\begin{equation}
D_{\mathrm{KL}}
\left(
\mathbf p
\|
\operatorname{softmax}
(\mathbf z+\mathbf v)
\right)
=
\frac{1}{2}
\mathbf v^{\top}
\mathbf F(\mathbf p)
\mathbf v
+
\mathcal O(\|\mathbf v\|^3).
\end{equation}

Hence, the Fisher energy provides a second-order approximation of the
task-level distribution shift induced by logit perturbations.

Based on this metric, LCTC separately measures the direct quantization defect
and trajectory-dependent sensitivity.
For the clean-state task defect $\mathbf a_t$ defined in
Appendix~\ref{app:lctc_transport}, the anchor loss is defined as

\begin{equation}
\mathcal L_{\mathrm{anchor}}
=
\mathbb E_{t,\mathbf x}
[
\mathcal E_{\mathrm F}
(
\mathbf a_t
)
].
\end{equation}

This objective encourages each loop-conditioned quantizer to minimize its
direct influence on the final prediction distribution.

For the trajectory secant $\mathbf s_t(\alpha)$, we define the trajectory
calibration loss as

\begin{equation}
\mathcal L_{\mathrm{traj}}
=
\mathbb E_{t,\mathbf x,\alpha}
[
\mathcal E_{\mathrm F}
(
\mathbf s_t(\alpha)
)
].
\end{equation}

Unlike $\mathcal L_{\mathrm{anchor}}$, which focuses on the clean recursive
state, $\mathcal L_{\mathrm{traj}}$ explicitly constrains the sensitivity of
quantization defects to inherited trajectory drift.

Since the two task-space losses may have different scales during
optimization, we normalize them before combining with other objectives:

\begin{equation}
\widetilde{\mathcal L}_{\mathrm{anchor}}
=
\frac{
\mathcal L_{\mathrm{anchor}}
}
{
\operatorname{stopgrad}
(
\mathcal L_{\mathrm{anchor}}
)
+\epsilon
},
\end{equation}

\begin{equation}
\widetilde{\mathcal L}_{\mathrm{traj}}
=
\frac{
\mathcal L_{\mathrm{traj}}
}
{
\operatorname{stopgrad}
(
\mathcal L_{\mathrm{traj}}
)
+\epsilon
}.
\end{equation}

The complete LCTC objective is formulated as

\begin{equation}
\mathcal L_{\mathrm{LCTC}}
=
\mathcal L_{\mathrm{KD}}
+
\lambda_{\mathrm{NLL}}
\mathcal L_{\mathrm{NLL}}
+
\lambda_{\mathrm a}
\widetilde{\mathcal L}_{\mathrm{anchor}}
+
\lambda_{\mathrm t}
\widetilde{\mathcal L}_{\mathrm{traj}}
+
\lambda_{\mathrm{reg}}
\mathcal R(\boldsymbol{\Gamma}).
\end{equation}

The regularization term constrains calibration parameters to remain close to
their initialization:

\begin{equation}
\mathcal R(\boldsymbol{\Gamma})
=
\sum_{m,t}
\left[
\|
\mathbf c_{m,t}
-
\mathbf c_{m,t}^{(0)}
\|_2^2
+
\|
\mathbf g_{m,t}
-
\mathbf 1
\|_2^2
+
\|
\boldsymbol{\beta}_{m,t}
\|_2^2
\right].
\end{equation}

The first term preserves the initial clipping behavior, the second term
regularizes gain correction around identity mapping, and the third term
prevents excessive bias compensation.

During optimization, all quantized weights, CLDC transformations, ILPA
precision assignments, and backbone parameters remain frozen.
Only the loop-conditioned calibration parameters
$\boldsymbol{\Gamma}_{m,t}$ are updated, making LCTC a lightweight
task-aligned calibration procedure rather than additional model training.

\subsection{Additional Experimental Details}
\label{app:experimental_details}

\subsubsection{CLDC Optimization Details}
\label{app:cldc_optimization}
For each quantized linear layer, CLDC learns a loop-shared diagonal scaling
and block-diagonal orthogonal transformation to construct a unified
quantization-friendly representation across recursive states. 
The scaling parameters are parameterized as logarithmic channel vectors, where
the actual scaling factors are obtained through exponential mapping and clipped
to $[10^{-3},10^{3}]$ for numerical stability. 
The orthogonal transformations are parameterized by block-diagonal Cayley maps
with a block size of 32. 
The Cayley parameters are initialized to zero, corresponding to identity
transformations, and incomplete final blocks are masked according to their
actual dimensions.

Both scaling and orthogonal transformation parameters are jointly optimized
using Adam with a learning rate of $1\times10^{-2}$ for 80 steps. 
No learning-rate scheduler is applied. 
Gradient norms are clipped to 1.0, and the parameters achieving the lowest
optimization objective are retained.

\subsubsection{LCTC Optimization Details}
\label{app:lctc_optimization}

LCTC optimizes only the loop-conditioned calibration parameters
$\boldsymbol{\Gamma}_{m,t}$ while keeping quantized weights, CLDC
transformations, ILPA precision assignments, and backbone parameters frozen.
The teacher-alignment term is implemented as the token-averaged KL divergence
between the teacher and student output distributions without temperature
scaling (i.e., $\tau_{\mathrm{KD}}=1$).

The interpolation coefficient $\alpha$ used in the trajectory-aware defect
estimation is sampled from $\{0.5,1.0,1.5\}$ with probabilities
$(0.25,0.50,0.25)$.
For the first loop, the trajectory term is disabled since no inherited
quantization drift exists.

The clipping parameters are initialized through local reconstruction search.
Specifically, for each module and loop, we select the clipping ratio from
$\{0.50,0.51,\ldots,1.00\}$ that minimizes the reconstruction error on
calibration activations.
The gain parameters are initialized to one and the bias parameters are
initialized to zero.

During calibration, the loss weights remain fixed:
$\lambda_{\mathrm{NLL}}=0.1$,
$\lambda_{\mathrm a}=0.1$,
$\lambda_{\mathrm t}=0.03$,
and
$\lambda_{\mathrm{reg}}=10^{-4}$.
The anchor and trajectory losses are normalized by their corresponding
pre-calibration reference values before being combined with other
optimization objectives.

\subsubsection{Deployment Storage Overhead}
\label{app:storage_overhead}

We further analyze the deployment storage overhead introduced by LoopQuant.
The analysis focuses on the additional persistent memory required by the
deployed model, including the standard W4 model storage and the static
metadata retained by CLDC, ILPA, and LCTC.
All calibration-time activation statistics, sensitivity scores, trajectory
caches, and optimization states are discarded after quantization and
therefore incur no deployment storage.

We first formulate the storage of the W4 base model.
Let $N_q$ denote the number of weights in the quantized linear layers,
$N_s$ the number of retained BF16 quantization scales, and $N_u$ the number
of parameters that remain unquantized in BF16, including embeddings, the LM
head, and normalization parameters.
The W4 base storage is
\begin{equation}
\mathcal{M}_{\mathrm{W4}}
=
\frac{N_q}{2}
+
2N_s
+
2N_u
\quad \text{bytes},
\label{eq:w4_storage}
\end{equation}
where the three terms correspond to packed 4-bit weights, BF16 quantization
scales, and unquantized BF16 parameters, respectively.
Thus, the actual W4 footprint is larger than the idealized model size
obtained by assigning 4 bits to every parameter, since quantization scales
and non-quantized parameters remain in BF16.

We next derive the additional storage introduced by LoopQuant.
Let $\mathcal{M}$ denote the set of transformed linear modules, $d_m$ the
channel dimension associated with module $m$, and
\begin{equation}
D_{\Sigma}
=
\sum_{m\in\mathcal{M}} d_m .
\end{equation}
For CLDC, each transformed module retains a BF16 diagonal scaling vector
$\mathbf{D}_m$, a BF16 block-diagonal rotation $\mathbf{R}_m$, and a
permutation vector $\mathbf{\Pi}_m$.
With a rotation block size $B$, a module of dimension $d_m$ contains $d_m/B$
blocks of size $B\times B$, resulting in $Bd_m$ stored rotation elements.
Summed over all transformed modules, the corresponding storage is
\begin{align}
\mathcal{M}_{D}
&=
2D_{\Sigma}, \\
\mathcal{M}_{R}
&=
2BD_{\Sigma}, \\
\mathcal{M}_{\Pi}
&=
8D_{\Sigma},
\end{align}
bytes, respectively, where $\mathbf{D}$ and $\mathbf{R}$ are stored in BF16
and $\mathbf{\Pi}$ uses 64-bit integer indices.
Therefore, the total CLDC overhead is
\begin{equation}
\mathcal{M}_{\mathrm{CLDC}}
=
(2B+10)D_{\Sigma}.
\label{eq:cldc_storage}
\end{equation}

ILPA only retains the final loop-wise activation-precision assignment.
If $N_{\mathrm{mask}}$ uint8 entries are used to encode the assignment, its
deployment storage is
\begin{equation}
\mathcal{M}_{\mathrm{ILPA}}
=
N_{\mathrm{mask}}
\quad \text{bytes}.
\label{eq:ilpa_storage}
\end{equation}
All sensitivity and interaction statistics used to obtain this assignment
are discarded after calibration.

For LCTC, each calibrated activation group retains three FP32 parameters
corresponding to clipping, gain, and bias.
Let $G_{m,t}$ denote the number of retained groups for module $m$ at
recurrent step $t$, and define
\begin{equation}
G_{\Sigma}
=
\sum_{m\in\mathcal{M}}
\sum_{t=1}^{T}
G_{m,t}.
\end{equation}
Since each group retains three FP32 scalars, the LCTC overhead is
\begin{equation}
\mathcal{M}_{\mathrm{LCTC}}
=
3\times4\times G_{\Sigma}
=
12G_{\Sigma}
\quad \text{bytes}.
\label{eq:lctc_storage}
\end{equation}

Combining the three components, the total additional deployment storage
introduced by LoopQuant is
\begin{equation}
\boxed{
\mathcal{M}_{\mathrm{extra}}
=
(2B+10)D_{\Sigma}
+
N_{\mathrm{mask}}
+
12G_{\Sigma}
}
\label{eq:loopquant_extra_storage}
\end{equation}
and the complete LoopQuant deployment footprint is
\begin{equation}
\boxed{
\mathcal{M}_{\mathrm{LoopQuant}}
=
\frac{N_q}{2}
+
2N_s
+
2N_u
+
(2B+10)D_{\Sigma}
+
N_{\mathrm{mask}}
+
12G_{\Sigma}
}.
\label{eq:loopquant_total_storage}
\end{equation}

Accordingly, the relative storage overhead over the standard W4 model is
\begin{equation}
\rho_{\mathrm{overhead}}
=
\frac{
(2B+10)D_{\Sigma}
+
N_{\mathrm{mask}}
+
12G_{\Sigma}
}{
\frac{N_q}{2}+2N_s+2N_u
}
\times100\%.
\label{eq:loopquant_overhead_ratio}
\end{equation}

The above formulation shows that the additional storage introduced by
LoopQuant scales only with the transformed channel dimensions, the number of
recurrent precision assignments, and the number of calibrated groups, rather
than with the full number of model weights.
Moreover, all retained quantities are static model metadata and require no
token-dependent cache or recurrent-history storage during inference.
The above analysis focuses on persistent storage overhead; the runtime cost
introduced by activation transformations and loop-conditioned calibration is
reflected in the end-to-end throughput evaluation in
Sec.~\ref{sec:efficiency}.
W4A4 and W4A6 use the same retained tensor shapes, differing only in the
loop-wise activation-precision assignment, and therefore incur essentially
identical deployment storage.

\subsubsection{Detailed Main Results}
\label{app:detailed main results}
Table~\ref{tab:app:loopquant_zero_shot_full} expands the averaged zero-shot results in Table~\ref{tab:main_results} into task-level accuracy on ARC-Easy, ARC-Challenge, HellaSwag, LAMBADA-openai, LAMBADA-standard, PIQA, and WinoGrande.
Under W4A6, LoopQuant achieves the highest average accuracy on all four LoopLMs despite using a lower average activation precision than the W4A8 baselines, outperforming the strongest W4A8 competitor by 1.93, 1.90, 0.64, and 1.36 percentage points on Ouro-1.4B, Ouro-2.6B, Nanbeige4.2-3B, and Huginn-0125, respectively.
The advantage becomes more pronounced under the challenging W4A4 setting, where LoopQuant improves the average accuracy over the strongest W4A4 baseline by 5.91, 5.31, 5.41, and 3.83 percentage points across the four models.
The task-level breakdown further shows that these gains are broadly distributed across benchmarks rather than arising from a single task; in particular, LoopQuant consistently achieves leading results across reasoning, language modeling, and completion tasks.
Notably, under W4A4, LoopQuant obtains the best quantized accuracy on all seven tasks across all four evaluated LoopLMs, demonstrating substantially stronger robustness to aggressive activation quantization.
These detailed results are consistent with the averaged accuracy and perplexity results in Table~\ref{tab:main_results}, further validating the effectiveness of LoopQuant for low-bit quantization under recursive computation.
\begin{table}[h]
\centering

\caption{Detailed zero-shot accuracy on ARC-Easy (AE), ARC-Challenge (AC), HellaSwag (HS), LAMBADA-openai (LO), LAMBADA-standard (LS), PIQA (PQ), and WinoGrande (WG). Bold and underline denote the best and second-best quantized results within each comparison group.}
\vspace{-0.3cm}
\resizebox{0.9\textwidth}{!}{
\tablestyle{2.5pt}{1.1}
\begin{tabular}{clccccccccc}
\toprule
\textbf{Model} & \textbf{Method} & \textbf{W/A}
& \textbf{AE}$\uparrow$ & \textbf{AC}$\uparrow$ & \textbf{HS}$\uparrow$
& \textbf{LO}$\uparrow$ & \textbf{LS}$\uparrow$
& \textbf{PQ}$\uparrow$ & \textbf{WG}$\uparrow$
& \textbf{Avg.}$^7$ $\uparrow$ \\
\midrule

\multirow{13}{*}{\makecell[l]{Ouro-1.4B}}
& BF16 & 16/16 & 76.81 & 52.13 & 71.59 & 65.17 & 57.75 & 78.51 & 65.98 & 66.85 \\
\cdashline{2-11}
& SmoothQuant & 4/8 & 69.42 & 46.21 & 66.33 & 54.69 & 48.67 & 75.31 & \underline{63.90} & 60.65 \\
& QuaRot & 4/8 & \underline{74.85} & \underline{50.65} & 67.55 & 56.51 & 51.39 & 74.28 & 62.48 & 62.53 \\
& SpinQuant & 4/8 & 73.59 & 48.00 & \underline{68.02} & \underline{58.99} & \underline{52.18} & \underline{75.48} & 62.56 & \underline{62.69} \\
& FlatQuant & 4/8 & 73.72 & 47.49 & 66.68 & 55.25 & 48.57 & 75.04 & 62.87 & 61.37 \\
& DartQuant & 4/8 & 71.27 & 46.21 & 65.44 & 56.02 & 47.74 & 75.20 & 61.77 & 60.52 \\
& \textbf{LoopQuant} & \textbf{4/6} & \textbf{77.51} & \textbf{51.69} & \textbf{68.75} & \textbf{60.22} & \textbf{53.10} & \textbf{76.38} & \textbf{64.67} & \textbf{64.62} \\
\cdashline{2-11}
& SmoothQuant & 4/4 & 65.28 & 40.37 & 61.10 & 45.29 & 38.91 & 70.16 & \underline{58.90} & 54.29 \\
& QuaRot & 4/4 & \underline{71.34} & \underline{45.49} & \underline{63.96} & \underline{50.98} & \underline{46.55} & 71.24 & 57.33 & \underline{58.13} \\
& SpinQuant & 4/4 & 69.19 & 44.29 & 62.02 & 47.08 & 40.52 & 70.65 & 56.62 & 55.77 \\
& FlatQuant & 4/4 & 70.07 & 45.15 & 62.46 & 49.91 & 43.76 & \underline{71.57} & 57.09 & 57.14 \\
& DartQuant & 4/4 & 61.49 & 37.47 & 57.59 & 43.64 & 37.78 & 67.71 & 54.56 & 51.46 \\

& \textbf{LoopQuant} & \textbf{4/4} & \textbf{77.33} & \textbf{51.42} & \textbf{68.94} & \textbf{59.36} & \textbf{52.26} & \textbf{75.58} & \textbf{63.37} & \textbf{64.04} \\
\midrule

\multirow{13}{*}{\makecell[l]{Ouro-2.6B}}
& BF16 & 16/16 & 80.93 & 58.53 & 76.52 & 69.98 & 62.97 & 80.14 & 68.75 & 71.12 \\
\cdashline{2-11}
& SmoothQuant & 4/8 & 77.21 & 55.00 & 72.75 & 63.20 & 57.98 & 77.92 & 65.80 & 67.12 \\
& QuaRot & 4/8 & \underline{78.22} & \underline{56.36} & \underline{73.45} & \underline{66.54} & \underline{59.67} & 77.98 & \underline{65.88} & \underline{68.30} \\
& SpinQuant & 4/8 & 75.65 & 53.63 & 73.40 & 64.80 & 57.64 & \underline{78.09} & 65.80 & 67.00 \\
& FlatQuant & 4/8 & 77.84 & 55.42 & 72.49 & 63.09 & 59.01 & 77.49 & 65.80 & 67.31 \\
& DartQuant & 4/8 & 77.84 & 54.57 & 71.35 & 59.67 & 54.22 & 76.67 & 65.25 & 65.65 \\

& \textbf{LoopQuant} & \textbf{4/6} & \textbf{80.29} & \textbf{57.45} & \textbf{76.25} & \textbf{68.04} & \textbf{60.97} & \textbf{80.16} & \textbf{68.26} & \textbf{70.20} \\
\cdashline{2-11}
& SmoothQuant & 4/4 & 72.43 & 48.30 & 68.25 & 56.76 & 51.46 & 73.69 & 61.43 & 61.76 \\
& QuaRot & 4/4 & 73.02 & \underline{51.46} & \underline{69.63} & \underline{60.54} & \underline{54.22} & \underline{74.56} & \underline{61.60} & \underline{63.58} \\
& SpinQuant & 4/4 & 70.49 & 46.17 & 66.09 & 50.71 & 46.30 & 72.66 & 60.17 & 58.94 \\
& FlatQuant & 4/4 & \underline{75.08} & 51.29 & 68.33 & 56.70 & 52.41 & 73.20 & 60.96 & 62.57 \\
& DartQuant & 4/4 & 68.31 & 43.01 & 62.72 & 48.44 & 40.19 & 69.83 & 57.33 & 55.69 \\

& \textbf{LoopQuant} & \textbf{4/4} & \textbf{79.50} & \textbf{56.09} & \textbf{75.07} & \textbf{65.88} & \textbf{60.35} & \textbf{78.58} & \textbf{66.75} & \textbf{68.89} \\
\midrule

\multirow{13}{*}{\makecell[l]{Nanbeige4.2-3B}}
& BF16 & 16/16 & 65.28 & 42.92 & 66.42 & 52.65 & 47.78 & 72.69 & 63.05 & 58.68 \\
\cdashline{2-11}
& SmoothQuant & 4/8 & \textbf{63.74} & \underline{42.37} & \underline{62.30} & \underline{49.78} & \textbf{44.32} & \underline{69.98} & 60.19 & \underline{56.10} \\
& QuaRot & 4/8 & 61.64 & 40.06 & 61.27 & 45.33 & 40.48 & 69.16 & 58.85 & 53.83 \\
& SpinQuant & 4/8 & 61.85 & 39.64 & 61.39 & 47.45 & 39.94 & 68.84 & \underline{62.17} & 54.47 \\
& FlatQuant & 4/8 & 61.64 & 40.15 & 62.26 & \textbf{50.09} & 44.01 & 69.49 & 58.37 & 55.14 \\
& DartQuant & 4/8 & 61.09 & 38.70 & 60.01 & 45.02 & 39.18 & 68.95 & 58.46 & 53.06 \\

& \textbf{LoopQuant} & \textbf{4/6} & \underline{62.45} & \textbf{42.82} & \textbf{63.95} & 49.70 & \underline{44.08} & \textbf{71.32} & \textbf{62.83} & \textbf{56.74} \\
\cdashline{2-11}
& SmoothQuant & 4/4 & \underline{59.47} & \underline{38.92} & 57.61 & \underline{43.20} & \underline{36.91} & 64.55 & 54.25 & \underline{50.70} \\
& QuaRot & 4/4 & 57.36 & 36.87 & 57.76 & 41.00 & 35.65 & \underline{64.66} & 54.88 & 49.74 \\
& SpinQuant & 4/4 & 53.03 & 34.57 & 57.05 & 39.90 & 33.92 & 64.28 & 55.59 & 48.33 \\
& FlatQuant & 4/4 & 53.70 & 35.76 & \underline{57.96} & 41.29 & 34.91 & 63.19 & \underline{57.17} & 49.14 \\
& DartQuant & 4/4 & 50.80 & 33.97 & 54.79 & 37.49 & 28.49 & 62.81 & 54.96 & 46.19 \\

& \textbf{LoopQuant} & \textbf{4/4} & \textbf{61.46} & \textbf{43.14} & \textbf{63.44} & \textbf{49.34} & \textbf{42.84} & \textbf{71.08} & \textbf{61.50} & \textbf{56.11} \\
\midrule

\multirow{13}{*}{\makecell[l]{Huginn-0125}}
& BF16 & 16/16 & 53.07 & 32.25 & 56.22 & 42.60 & 39.88 & 69.48 & 55.72 & 49.89 \\
\cdashline{2-11}
& SmoothQuant & 4/8 & 49.89 & 29.06 & 53.31 & 38.99 & 37.26 & 66.72 & 53.41 & 46.95 \\
& QuaRot & 4/8 & 50.61 & 30.25 & 53.60 & \underline{40.44} & \underline{37.76} & 67.21 & \underline{54.27} & \underline{47.73} \\
& SpinQuant & 4/8 & 50.15 & 29.83 & \underline{53.68} & 40.19 & 37.05 & 67.04 & 53.40 & 47.33 \\
& FlatQuant & 4/8 & 49.98 & 30.08 & 53.49 & 39.90 & 37.16 & 66.82 & 53.72 & 47.31 \\
& DartQuant & 4/8 & \underline{50.86} & \underline{30.34} & 53.51 & 39.70 & 36.15 & \underline{67.37} & 53.33 & 47.32 \\

& \textbf{LoopQuant} & \textbf{4/6} & \textbf{52.38} & \textbf{31.38} & \textbf{54.92} & \textbf{42.25} & \textbf{38.60} & \textbf{69.72} & \textbf{54.35} & \textbf{49.09} \\
\cdashline{2-11}
& SmoothQuant & 4/4 & 46.59 & 26.72 & 50.44 & 36.64 & 34.13 & 63.57 & 50.62 & 44.10 \\
& QuaRot & 4/4 & \underline{47.98} & 27.14 & \underline{51.06} & 36.81 & 34.37 & 64.23 & 50.14 & 44.53 \\
& SpinQuant & 4/4 & 46.67 & \underline{28.34} & 49.99 & \underline{37.04} & \underline{34.40} & 63.90 & 51.17 & 44.50 \\
& FlatQuant & 4/4 & 47.14 & 27.48 & 50.22 & 36.09 & 33.98 & \underline{64.93} & \underline{52.05} & \underline{44.56} \\
& DartQuant & 4/4 & 47.14 & 27.23 & 50.55 & 36.99 & 33.40 & 63.52 & 49.59 & 44.06 \\

& \textbf{LoopQuant} & \textbf{4/4} & \textbf{51.70} & \textbf{30.43} & \textbf{54.48} & \textbf{40.41} & \textbf{37.60} & \textbf{68.58} & \textbf{55.53} & \textbf{48.39} \\

\bottomrule
\end{tabular}}
\vspace{-0.2cm}
\label{tab:app:loopquant_zero_shot_full}
\end{table}

\subsubsection{Comparison with LoopQ}
\label{app:loopq_comparison}

LoopQ~\cite{fang2026loopq} is the first work that studies low-bit
quantization for recursive Transformer architectures, making it the closest
related approach to LoopQuant. Both methods recognize that recursive
computation introduces quantization challenges beyond conventional
feed-forward Transformers. However, they focus on different aspects of the
problem.

LoopQ mainly investigates cross-loop distribution shifts and recursive
quantization errors, and improves quantization robustness through loop-aware
calibration. In contrast, LoopQuant develops a more comprehensive framework
that jointly considers three key challenges in LoopLM quantization:
cross-loop activation dynamics, heterogeneous loop sensitivity, and the
misalignment between local quantization objectives and recursive task
behavior. Specifically, CLDC learns a shared quantization-friendly activation
geometry across recurrent states, ILPA performs loop-wise activation
precision allocation by modeling task sensitivity and cross-loop
interactions, and LCTC further calibrates quantization behavior along the
low-bit recursive trajectory.

To further compare with LoopQ, we report the results originally published in
the LoopQ paper and compare them with our corresponding results.
Since LoopQ does not provide publicly available implementation and complete
experimental configurations, we cannot perform a fully controlled
reproduction. Therefore, we only compare the overlapping models and
evaluation metrics reported by both works.

Table~\ref{tab:loopq_comparison} presents the comparison between LoopQ and
LoopQuant under different low-bit quantization settings. Under the more
challenging W4A4 setting, LoopQuant consistently achieves better robustness
on both Ouro-1.4B and Ouro-2.6B. For example, on Ouro-2.6B, LoopQuant improves
the four zero-shot metrics from 68.60/62.27/56.80/50.34 to
75.07/66.75/65.88/56.09, while reducing WikiText2 perplexity from 15.89 to
8.27. These results indicate that LoopQuant better preserves model
capability under extremely low activation precision.

Beyond the quantitative comparison, the two methods adopt different
optimization perspectives. LoopQ mainly focuses on mitigating distribution
shifts introduced by recursive execution, whereas LoopQuant further models
loop-wise precision sensitivity and task-level error propagation along the
complete recursive trajectory. By jointly optimizing activation geometry,
precision allocation, and trajectory-aware calibration, LoopQuant provides a
more comprehensive solution for low-bit LoopLM deployment.

\begin{table}[t]
\centering
\caption{
Comparison with LoopQ under different low-bit quantization settings.
LoopQ results are directly taken from its original paper due to the lack of
public implementation. We only compare overlapping models and evaluation
metrics.
}
\label{tab:loopq_comparison}
\resizebox{0.7\linewidth}{!}{
\begin{tabular}{lccccccc}
\toprule
Model & W/A & Method &
HS$\uparrow$ &
WG$\uparrow$ &
LB$\uparrow$ &
AC$\uparrow$ &
WT2$\downarrow$ \\
\midrule

\multirow{4}{*}{Ouro-1.4B}
& W4A8 & LoopQ
& \textbf{69.54}
& 64.56
& \textbf{62.55}
& 51.45
& 13.29
\\

& W4A6 & LoopQuant
& 68.75
& \textbf{64.67}
& 60.22
& \textbf{51.69}
& \textbf{9.48}
\\

& W4A4 & LoopQ
& 65.82
& 59.51
& 56.59
& 48.81
& 15.81
\\

& W4A4 & LoopQuant
& \textbf{68.94}
& \textbf{63.37}
& \textbf{59.36}
& \textbf{51.42}
& \textbf{9.86}
\\

\midrule

\multirow{4}{*}{Ouro-2.6B}
& W4A8 & LoopQ
& 75.55
& 67.80
& \textbf{68.37}
& \textbf{58.19}
& 11.38
\\

& W4A6 & LoopQuant
& \textbf{76.25}
& \textbf{68.26}
& 68.04
& 57.45
& \textbf{8.04}
\\

& W4A4 & LoopQ
& 68.60
& 62.27
& 56.80
& 50.34
& 15.89
\\

& W4A4 & LoopQuant
& \textbf{75.07}
& \textbf{66.75}
& \textbf{65.88}
& \textbf{56.09}
& \textbf{8.27}
\\

\bottomrule
\end{tabular}
}
\end{table}

\subsubsection{Ablation on Calibration Data}
\label{app:calibration_ablation}

\begin{figure*}[t]
    \centering
    \includegraphics[width=0.92\linewidth]{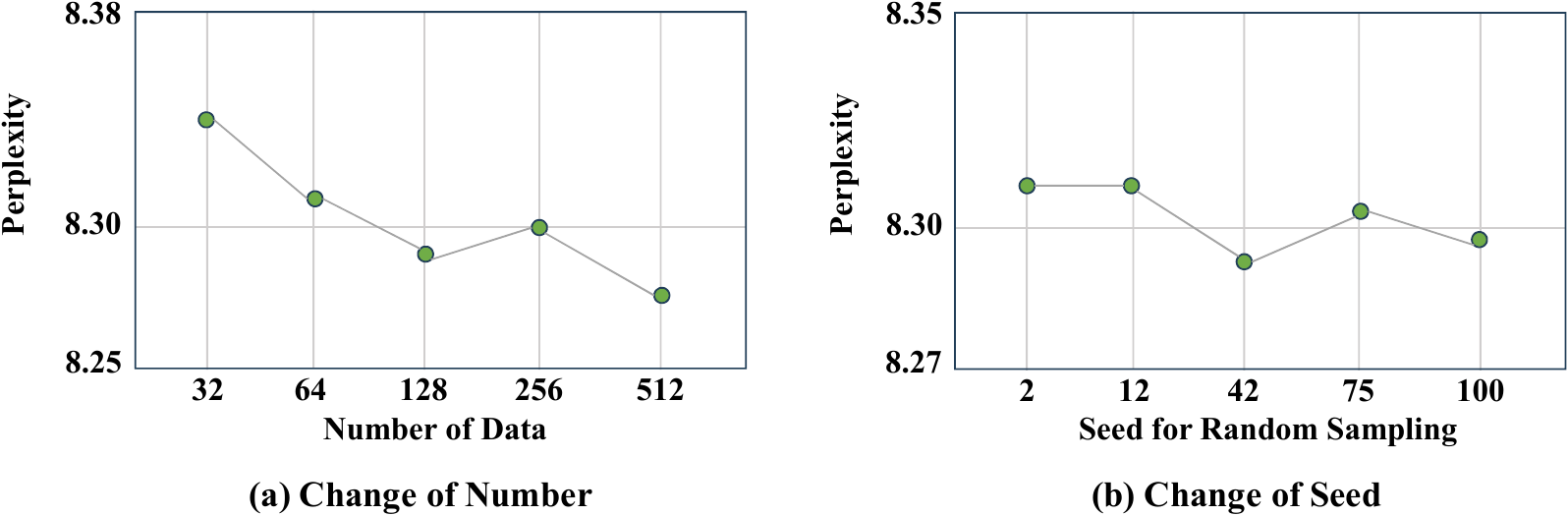}
    \caption{Sensitivity of LoopQuant to calibration-data configurations. 
(a) WikiText2 perplexity with different numbers of calibration samples.
(b) WikiText2 perplexity under different random sampling seeds.
All other quantization hyperparameters are kept unchanged.}
\label{fig:calibration_ablation}
\end{figure*}

We first investigate the sensitivity of LoopQuant to calibration-set size and sampling randomness.
Following the default calibration protocol, we vary the number of calibration samples over $\{32, 64, 128, 256, 512\}$ while keeping all other quantization hyperparameters unchanged.
For the seed-sensitivity study, we fix the calibration-set size to the default setting and vary the random sampling seed over $\{2, 12, 42, 75, 100\}$.
As shown in Fig.~\ref{fig:calibration_ablation}(a), LoopQuant remains stable across different calibration sizes, with a perplexity variation of less than 0.05.
The relationship is not strictly monotonic: increasing the number of samples generally provides only marginal gains, and the difference between 256 and 512 samples is approximately 0.02 in perplexity.
Similarly, Fig.~\ref{fig:calibration_ablation}(b) shows only minor fluctuations across different sampling seeds, with a maximum difference of approximately 0.02.
These results indicate that LoopQuant is insensitive to both calibration-set size and sampling randomness, suggesting that a modest calibration set is sufficient to obtain stable quantization performance.

\begin{table}[h]
\centering
\caption{Ablation on the choice of calibration corpus under W4A4. 
We compare WikiText2 and C4 calibration data on Ouro-1.4B and Ouro-2.6B.
Bold and underline denote the best and second-best results within each model, respectively.}
\resizebox{0.62\textwidth}{!}{
\tablestyle{2.5pt}{1.1}
\begin{tabular}{lcccc}
\toprule
\textbf{Model} & \textbf{Calibration Data}
& \textbf{WikiText2}$\downarrow$
& \textbf{C4}$\downarrow$
& \textbf{Avg. Acc.}$\uparrow$ \\
\midrule

\multirow{2}{*}{Ouro-1.4B}
& WikiText2
& \textbf{9.86}
& \underline{14.00}
& \textbf{64.04} \\
& C4
& \underline{10.57}
& \textbf{11.86}
& \underline{63.79} \\
\midrule

\multirow{2}{*}{Ouro-2.6B}
& WikiText2
& \textbf{8.27}
& \underline{12.24}
& \textbf{68.89} \\
& C4
& \underline{9.41}
& \textbf{10.15}
& \underline{68.07} \\

\bottomrule
\end{tabular}}
\vspace{-0.2cm}
\label{tab:app:calibration_dataset}
\end{table}
We further study the effect of calibration-corpus choice by calibrating LoopQuant on either WikiText2 or C4 under W4A4.
As summarized in Table~\ref{tab:app:calibration_dataset}, the choice of calibration corpus has only a moderate impact on the averaged zero-shot accuracy, with WikiText2 calibration achieving slightly higher accuracy on both Ouro-1.4B and Ouro-2.6B.
In contrast, perplexity exhibits a clear domain-matching effect.
Calibrating on WikiText2 yields lower WikiText2 perplexity, reducing it from 10.57 to 9.86 on Ouro-1.4B and from 9.41 to 8.27 on Ouro-2.6B compared with C4 calibration.
Conversely, C4 calibration substantially improves C4 perplexity, from 14.00 to 11.86 and from 12.24 to 10.15 on the two models, respectively.
These results suggest that the calibration corpus primarily affects the reconstruction quality of the corresponding evaluation domain, while LoopQuant maintains relatively stable downstream zero-shot accuracy across different calibration sources.

\subsubsection{Effectiveness of ILPA with Different Quantization Backbones}
\label{app:ilpa_backbone}

To further isolate the contribution of ILPA from the complete LoopQuant
pipeline, we apply the loop-wise precision allocation generated by ILPA to
different quantization backbones on Huginn-0125 ($T=6$). Specifically, we
replace the original uniform activation precision setting with the
ILPA-selected allocation while keeping other components of each backbone
unchanged. All methods are evaluated under the same average activation
precision budget of 6 bits.

As shown in Table~\ref{tab:ilpa_backbone}, applying ILPA consistently
improves different quantization backbones under the same activation budget.
For QuaRot, ILPA reduces WikiText2 perplexity from 14.02 to 13.59 and
improves the average zero-shot accuracy from 46.28\% to 47.35\%.
Similarly, SpinQuant benefits from the ILPA allocation, improving the
average accuracy from 45.97\% to 47.04\%. These results demonstrate that the
performance gains of ILPA come from task-aware loop-wise precision
allocation rather than simply using a mixed-precision configuration.
Combined with CLDC and LCTC, LoopQuant further achieves 12.47 WikiText2
perplexity and 49.09\% average accuracy, demonstrating the complementary
effects of precision allocation and trajectory-aware calibration.

\begin{table}[t]
\centering
\caption{
Effectiveness of ILPA with different quantization backbones on
Huginn-0125 ($T=6$) under the same average activation precision budget of
6 bits.
}
\label{tab:ilpa_backbone}
\begin{tabular}{lccc}
\toprule
Method & Allocation & Wiki2$\downarrow$ & Avg. Acc.$\uparrow$ \\
\midrule
QuaRot 
& Uniform A6 
& 14.02 
& 46.28
\\
QuaRot + ILPA
& [4,4,4,8,8,8]
& 13.59
& 47.35
\\
SpinQuant 
& Uniform A6
& 14.21
& 45.97
\\
SpinQuant + ILPA
& [4,4,4,8,8,8]
& 13.63
& 47.04
\\
LoopQuant
& ILPA + CLDC + LCTC
& \textbf{12.47}
& \textbf{49.09}
\\
\bottomrule
\end{tabular}
\end{table}

\subsubsection{Ablation on LCTC Calibration Parameters}
\label{app:lctc_parameter_ablation}

\begin{table}[h]
\centering
\caption{Ablation of the three calibration parameters in LCTC under W4A6.
We progressively introduce clipping, gain, and bias while keeping CLDC and ILPA fixed.
The storage overhead is measured relative to the W4 base model.}
\resizebox{0.72\textwidth}{!}{
\tablestyle{2.5pt}{1.1}
\begin{tabular}{c ccc c cc}
\toprule
\multirow{2}{*}{\textbf{W/A}}
& \multicolumn{3}{c}{\textbf{LCTC}}
& \multirow{2}{*}{\textbf{Storage Overhead}}
& \multirow{2}{*}{\textbf{WikiText2}$\downarrow$}
& \multirow{2}{*}{\textbf{C4}$\downarrow$ }\\
\cmidrule(lr){2-4}
& $\mathbf{c}_{m,t}$
& $\mathbf{g}_{m,t}$
& $\boldsymbol{\beta}_{m,t}$
& & & \\
\midrule

\multirow{4}{*}{W4A6}
& $\times$ & $\times$ & $\times$
& 0
& 27.05 & 34.17 \\

& $\checkmark$ & $\times$ & $\times$
& +0.49 MB (0.03\%)
& 26.48 & 33.38 \\

& $\checkmark$ & $\checkmark$ & $\times$
& +0.97 MB (0.05\%)
& 26.09 & 32.74 \\

& $\checkmark$ & $\checkmark$ & $\checkmark$
& +1.46 MB (0.08\%)
& \textbf{25.84} & \textbf{32.19} \\

\bottomrule
\end{tabular}}
\vspace{-0.2cm}
\label{tab:lctc_component_ablation}
\end{table}
We further ablate the three learnable calibration parameters in LCTC, namely clipping $\mathbf{c}_{m,t}$, gain $\mathbf{g}_{m,t}$, and bias $\boldsymbol{\beta}_{m,t}$, while keeping CLDC and ILPA fixed.
As shown in Table~\ref{tab:lctc_component_ablation}, removing LCTC entirely results in WikiText2 and C4 perplexities of 27.05 and 34.17, respectively.
Introducing clipping alone substantially improves the reconstruction quality, reducing the two perplexities to 26.48 and 33.38.
Adding gain further improves them to 26.09 and 32.74, while incorporating bias yields the full LCTC performance of 25.84 and 32.19.
Overall, the complete parameterization reduces WikiText2 and C4 perplexity by 1.21 and 1.98, respectively, compared with the variant without LCTC.
These results indicate that the three parameters provide complementary calibration capabilities: clipping controls the effective activation range, while gain and bias further correct the scale and distributional offset accumulated during recursive computation.
Importantly, these improvements require only 1.46\,MB of additional storage, corresponding to approximately 0.08\% of the W4 base-model footprint, demonstrating that LCTC provides consistent performance gains with negligible deployment overhead.

\subsubsection{Layer-wise Quantization Error during Reasoning}
\label{app:layerwise_error}

\begin{figure}[h]
    \centering

    \includegraphics[width=0.9\textwidth]{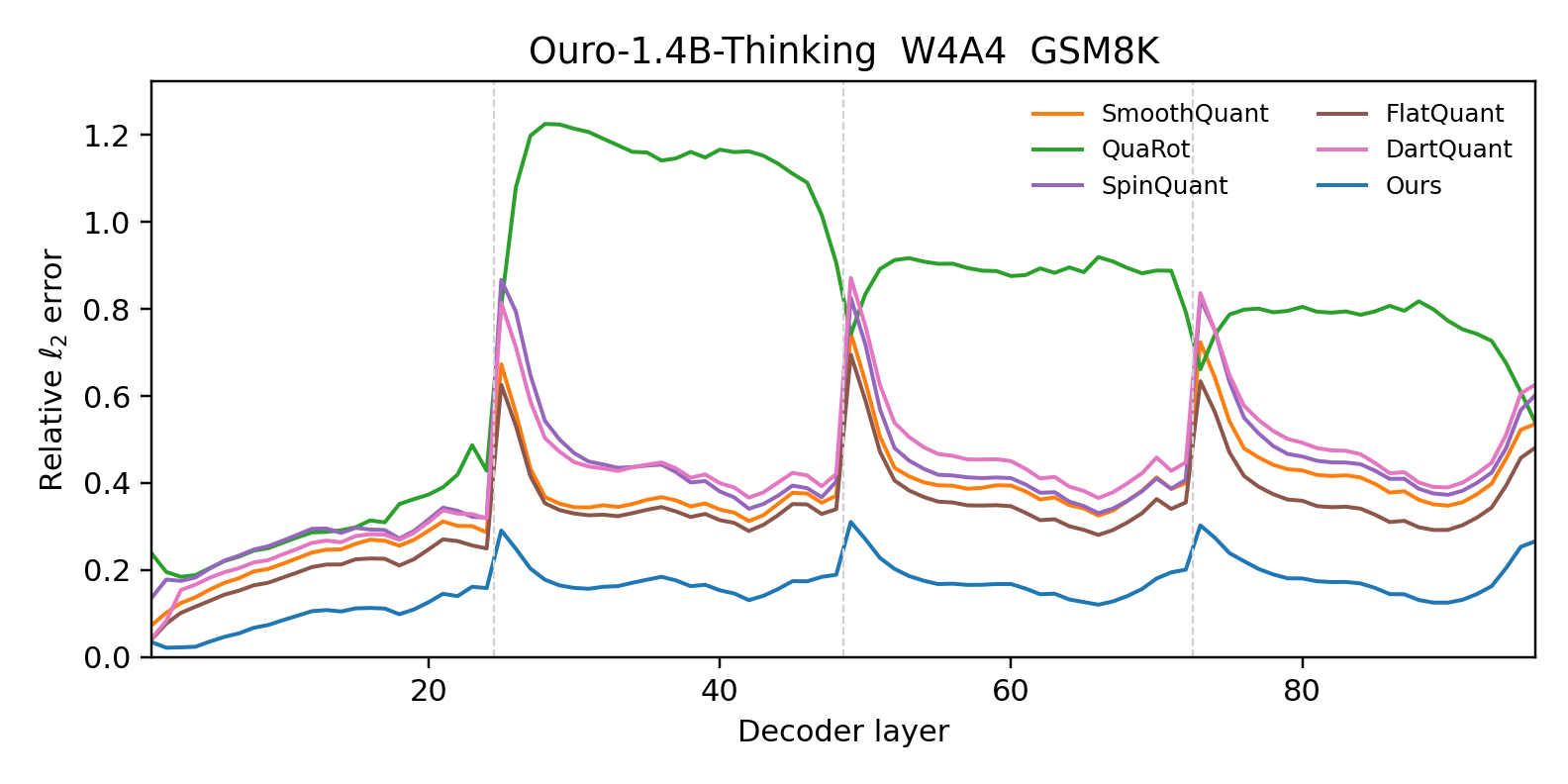}

    \vspace{0.15cm}

    \includegraphics[width=0.9\textwidth]{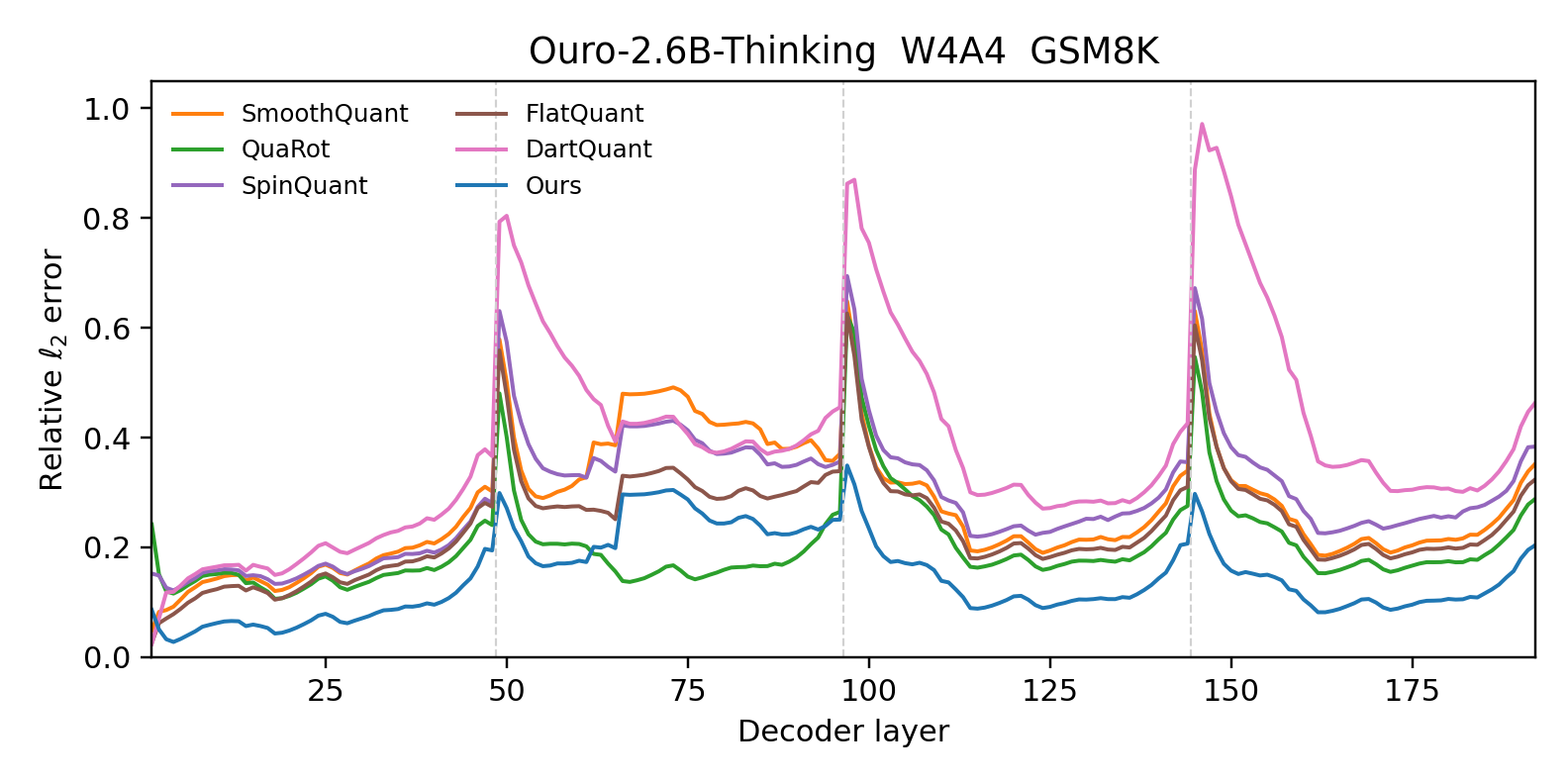}

    \caption{
    Layer-wise relative $\ell_2$ representation error on GSM8K under W4A4 quantization for Ouro-1.4B-thinking (top) and Ouro-2.6B-thinking (bottom).
    The error is measured against the corresponding BF16 hidden representation at each decoder layer.
    Dashed vertical lines indicate the boundaries between successive recurrent passes in the unrolled reasoning computation.
    }
    \label{fig:layerwise_error}
\end{figure}

To further understand how quantization errors propagate through recursive reasoning, we analyze the layer-wise representation deviation of different W4A4 methods on GSM8K.
For each evaluated reasoning trace, we run the BF16 and quantized models with the same input and record the hidden representation after each decoder layer.
Specifically, for decoder layer $l$, we measure the relative $\ell_2$ error with respect to the BF16 reference as
\begin{equation}
    e_l =
    \frac{
        \left\|
        \widehat{\mathbf{H}}_l-\mathbf{H}_l
        \right\|_2
    }{
        \left\|
        \mathbf{H}_l
        \right\|_2
    },
\end{equation}
where $\mathbf{H}_l$ and $\widehat{\mathbf{H}}_l$ denote the BF16 and quantized hidden representations, respectively.
We report the mean error collected over the evaluated GSM8K reasoning traces.
The dashed vertical lines in Fig.~\ref{fig:layerwise_error} indicate the boundaries between successive recurrent passes in the unrolled computation.

As shown in Fig.~\ref{fig:layerwise_error}, conventional PTQ methods exhibit pronounced error accumulation and periodic error spikes throughout recursive reasoning.
These spikes appear particularly around the beginning of a new recurrent pass, indicating that quantization perturbations inherited from the preceding computation can be re-amplified when the shared decoder is repeatedly applied.
This effect is especially evident on Ouro-1.4B-thinking: for example, QuaRot exhibits a sharp increase in relative error after the first recurrent boundary, with the deviation exceeding $1.2$ over several subsequent layers.
Other baselines, including SmoothQuant, SpinQuant, FlatQuant, and DartQuant, also show repeated error surges at later recurrent boundaries.

In contrast, LoopQuant substantially suppresses both the overall representation deviation and the magnitude of these periodic error spikes.
On Ouro-1.4B-thinking, its relative error remains markedly below the competing methods throughout nearly the entire reasoning trajectory, including immediately after recurrent transitions where other methods deteriorate most severely.
A similar trend can be observed on Ouro-2.6B-thinking.
Although some baselines remain competitive within individual local regions, LoopQuant consistently limits the peak error around recurrent boundaries and rapidly returns to a lower-error trajectory after each transition.
In particular, the error peaks of several baselines approach approximately $0.6$--$1.0$, whereas LoopQuant keeps the corresponding peaks substantially lower.

These observations provide direct evidence for the recursive error-accumulation problem studied in this work.
Unlike a conventional feed-forward Transformer, a LoopLM repeatedly applies shared parameters to hidden states that have already been perturbed by preceding quantized computation.
Consequently, even moderate local quantization errors can be repeatedly inherited, transformed, and amplified across recurrent passes.
The substantially flatter error trajectory of LoopQuant suggests that its loop-aware calibration and precision allocation more effectively preserve the intermediate representations encountered during recursive inference.
This layer-wise behavior is also consistent with the stronger W4A4 reasoning accuracy reported in the main results, showing that the benefit of LoopQuant arises not only from reducing local reconstruction error, but also from preventing quantization perturbations from accumulating along the reasoning trajectory.

\subsubsection{Activation Distribution Transformation by CLDC}
\label{app:cldc_distribution}

\begin{figure}[p]
    \centering

    \includegraphics[
        width=\textwidth,
        height=0.45\textheight,
        keepaspectratio
    ]{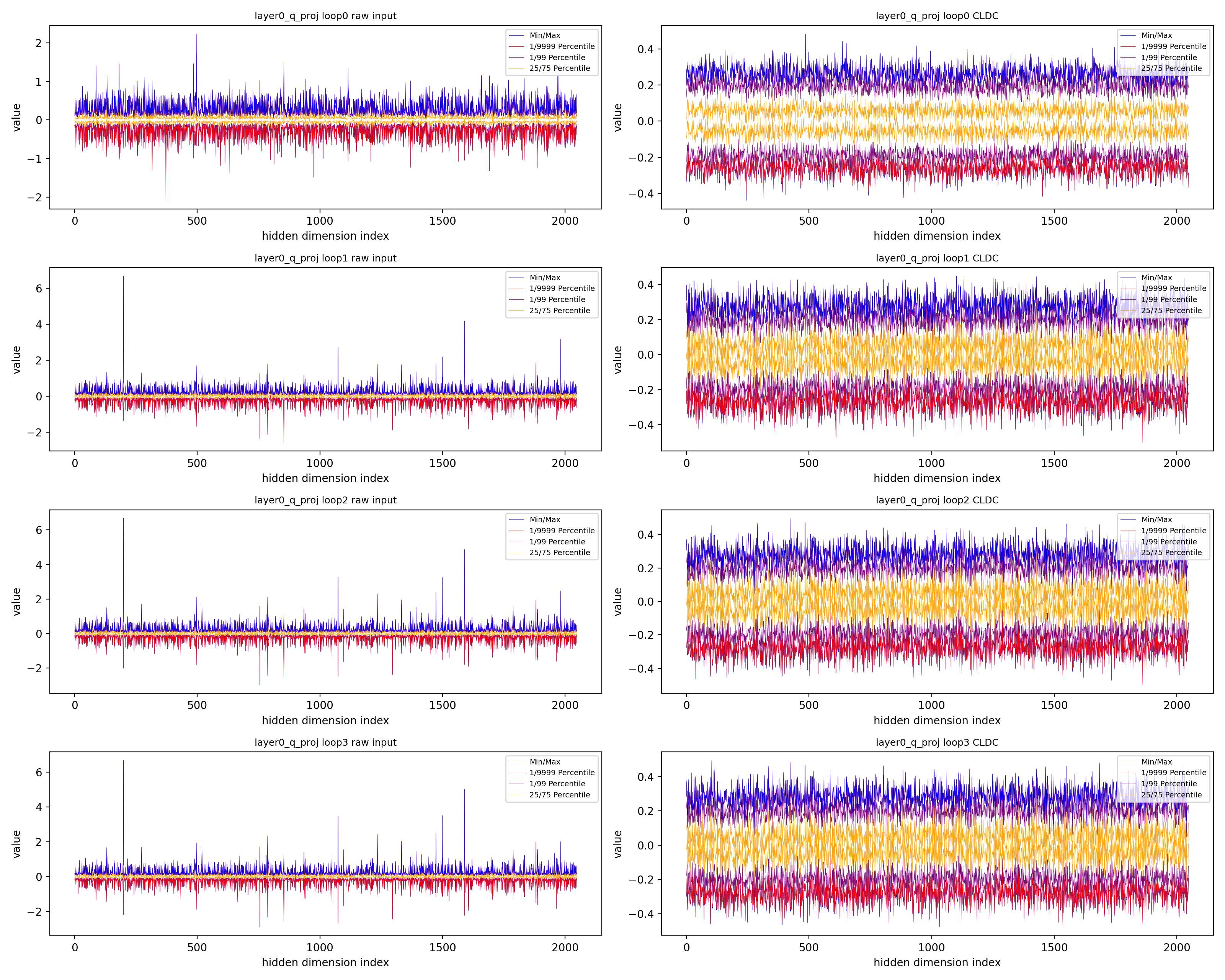}

    \vspace{0.3cm}

    \includegraphics[
        width=\textwidth,
        height=0.45\textheight,
        keepaspectratio
    ]{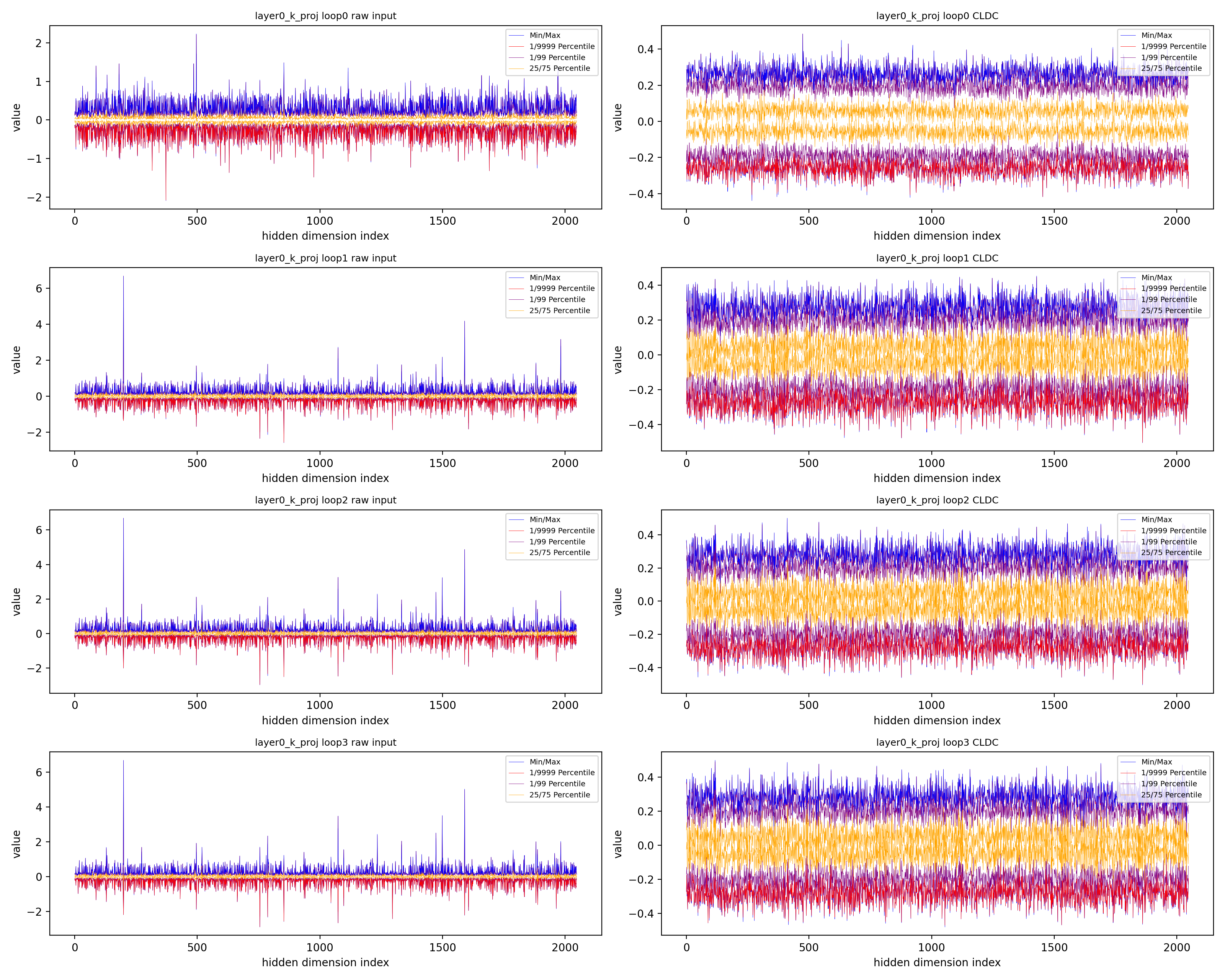}

    \caption{
    Channel-wise input activation distributions of the $q$ projection (top) and $k$ projection (bottom) in the first recurrent layer of Ouro-1.4B.
    Within each image, the left and right columns show the raw input and the corresponding representation after CLDC, respectively, while the four rows correspond to different recurrent loops.
    }
    \label{fig:cldc_dist_qk}
\end{figure}

\begin{figure}[p]
    \centering

    \includegraphics[
        width=\textwidth,
        height=0.45\textheight,
        keepaspectratio
    ]{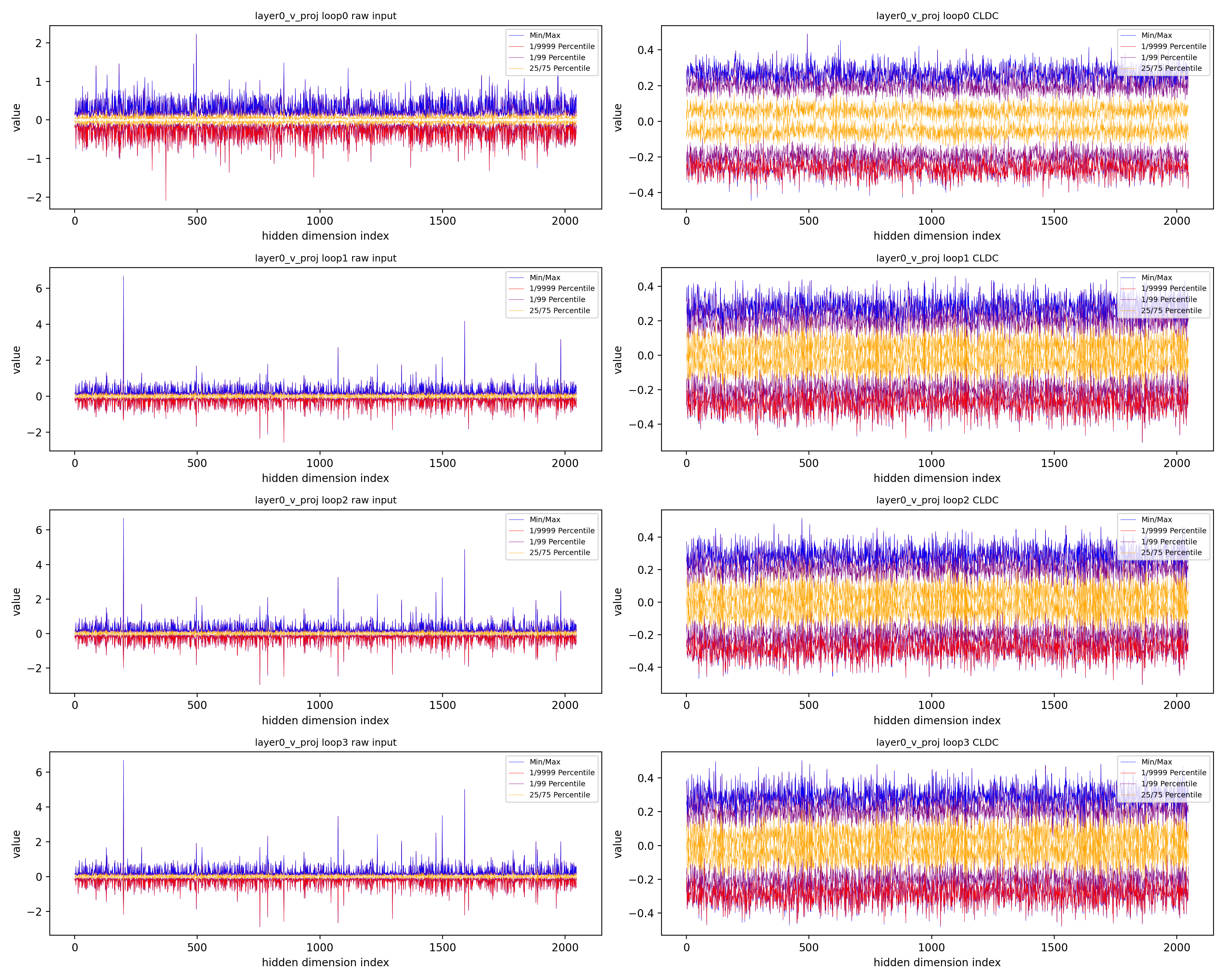}

    \vspace{0.3cm}

    \includegraphics[
        width=\textwidth,
        height=0.45\textheight,
        keepaspectratio
    ]{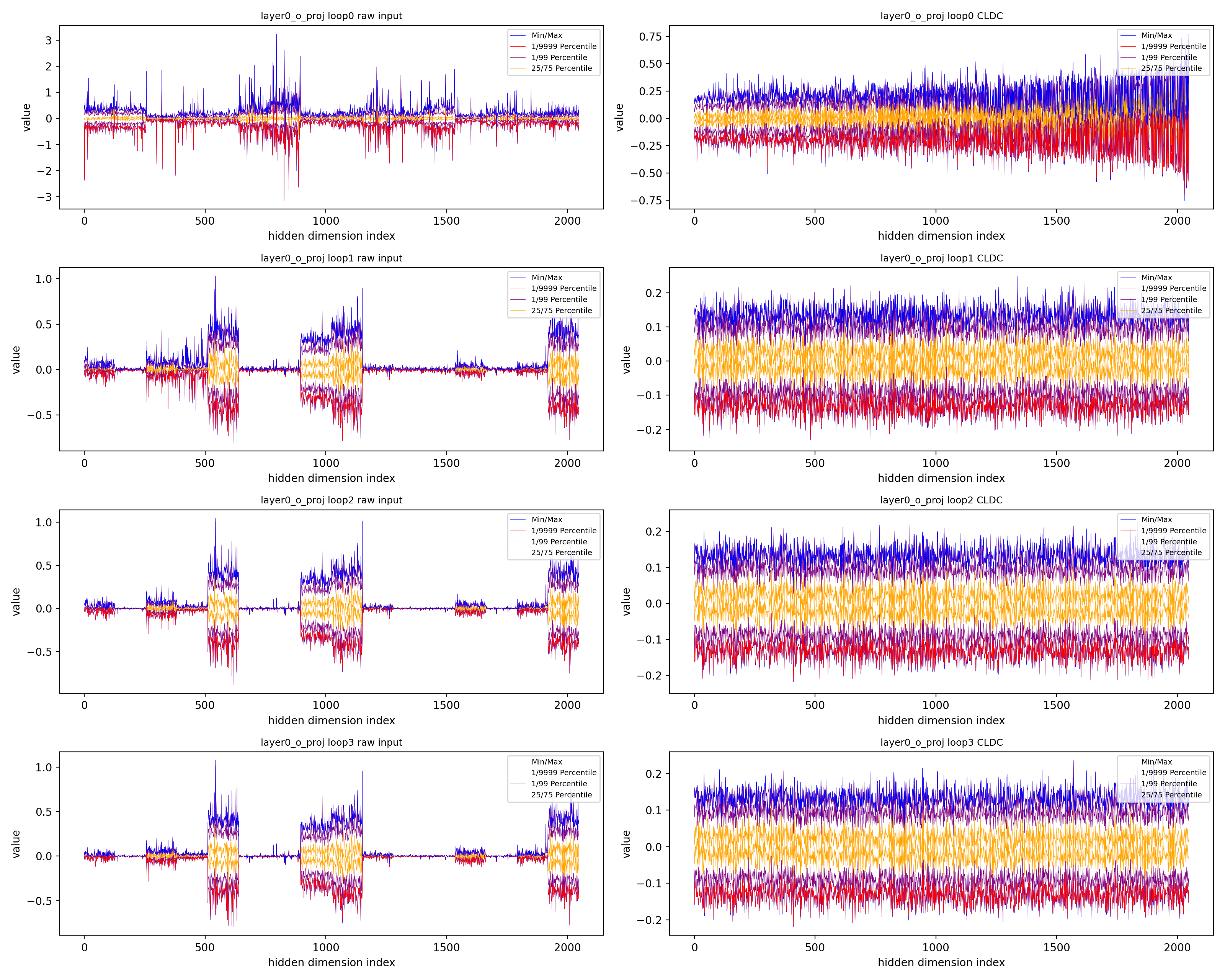}

    \caption{
    Channel-wise input activation distributions of the $v$ projection (top) and $o$ projection (bottom) in the first recurrent layer of Ouro-1.4B, before and after CLDC.
    The four rows correspond to different recurrent loops.
    CLDC substantially suppresses channel-wise outliers and produces more balanced activation ranges across hidden dimensions.
    }
    \label{fig:cldc_dist_vo}
\end{figure}

\begin{figure}[p]
    \centering

    \includegraphics[
        width=\textwidth,
        height=0.45\textheight,
        keepaspectratio
    ]{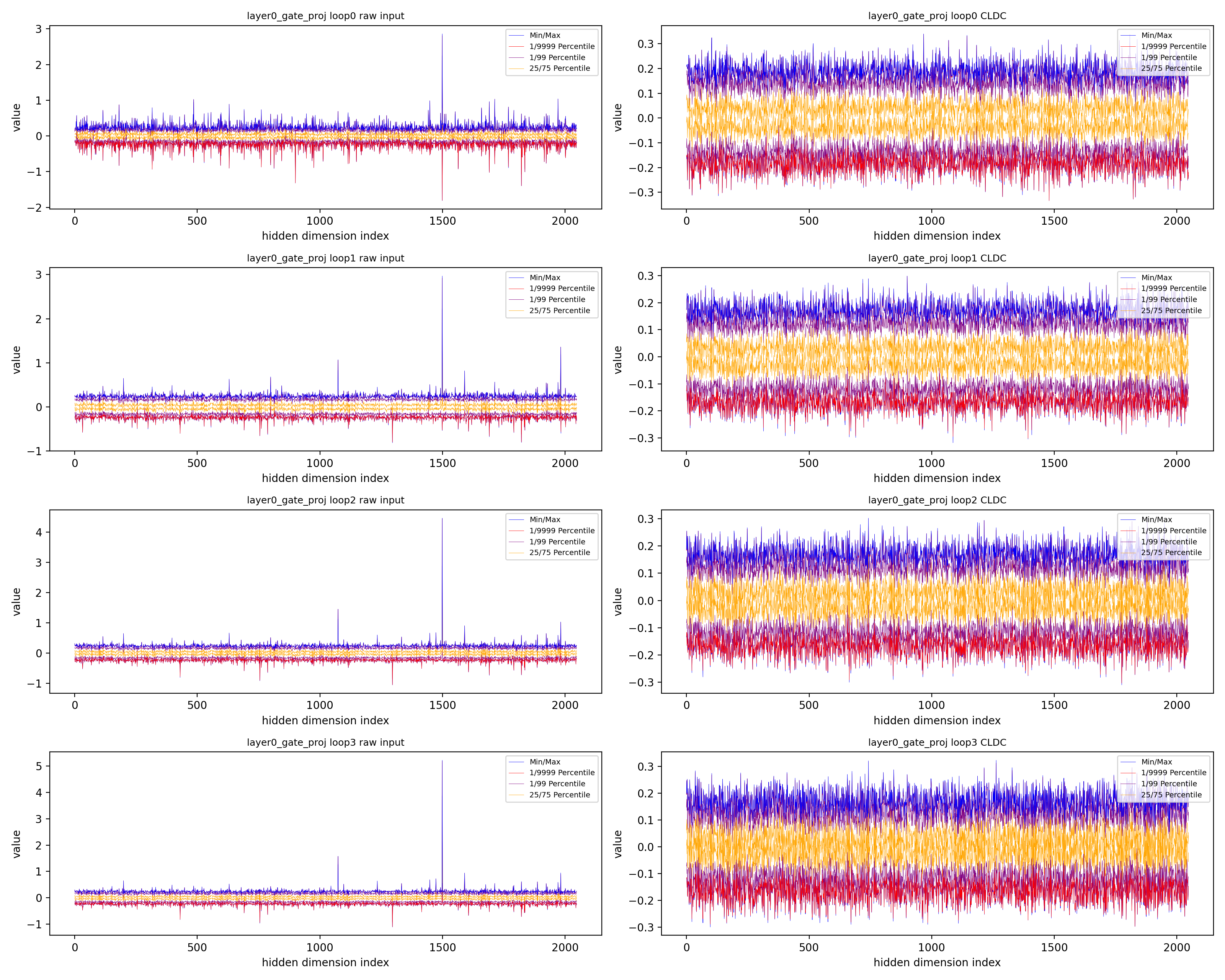}

    \vspace{0.3cm}

    \includegraphics[
        width=\textwidth,
        height=0.45\textheight,
        keepaspectratio
    ]{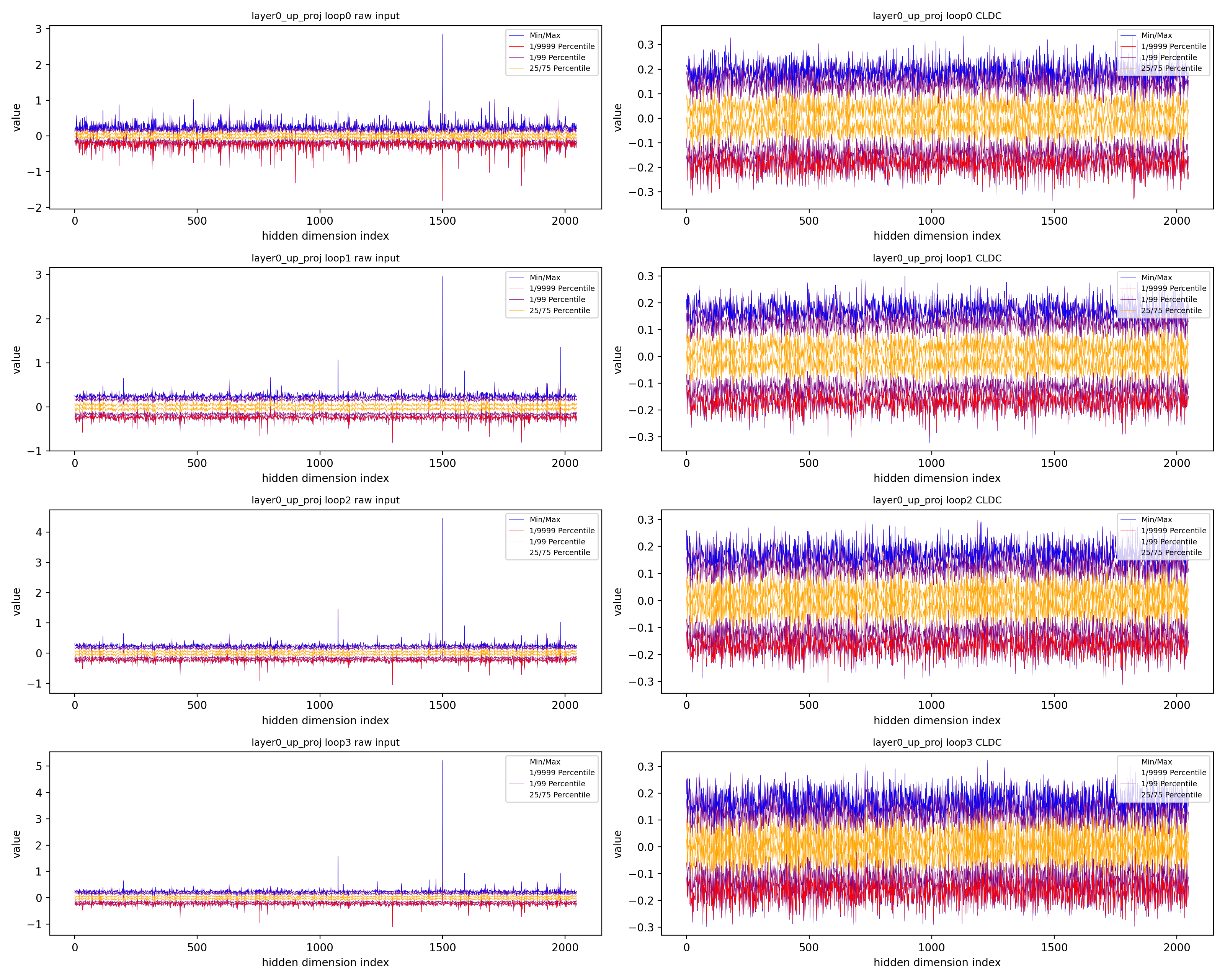}

    \caption{
    Channel-wise input activation distributions of the MLP gate projection (top) and up projection (bottom) in the first recurrent layer of Ouro-1.4B.
    Raw activations exhibit sparse but large channel-wise outliers across recurrent loops, whereas CLDC produces substantially more compact and homogeneous percentile envelopes.
    }
    \label{fig:cldc_dist_mlp}
\end{figure}

\begin{figure}[p]
    \centering

    \includegraphics[
        width=0.92\textwidth
    ]{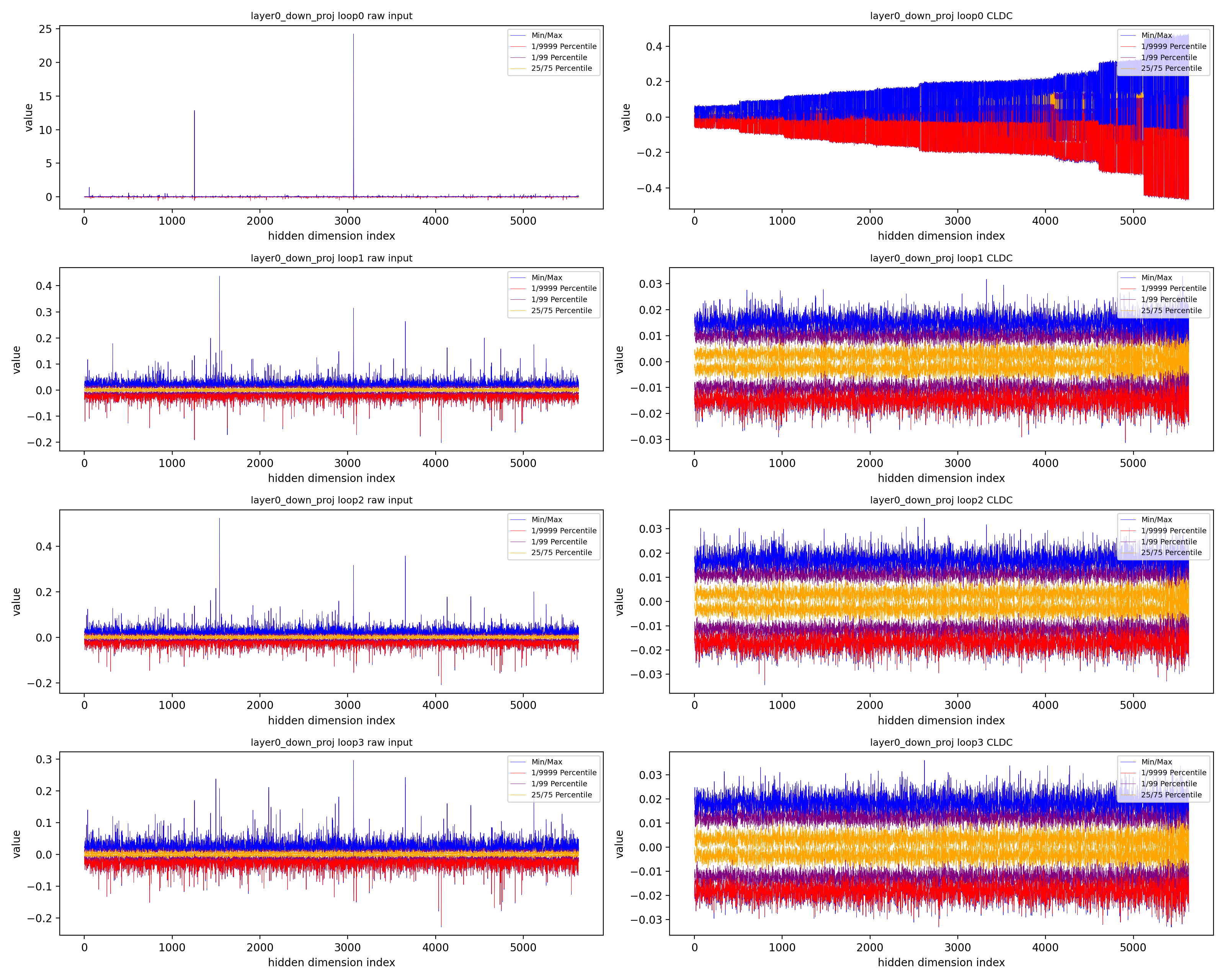}

    \caption{
    Channel-wise input activation distributions of the MLP down projection in the first recurrent layer of Ouro-1.4B.
    The left and right columns show the raw input and the representation after CLDC, respectively, across four recurrent loops.
    The raw input contains highly concentrated channel-wise outliers, including extreme values above 20 in the first loop, while most percentile statistics remain close to zero.
    CLDC strongly suppresses these extreme channels and transforms the activation space into a substantially more compact and balanced distribution.
    }
    \label{fig:cldc_dist_down}
\end{figure}

To provide a more direct view of how CLDC reshapes the activation space before low-bit quantization, we visualize the input distributions of different linear modules in the first recurrent layer of Ouro-1.4B.
For each module and loop, we collect its input activations over sampled sequences and compute channel-wise statistics across all token positions.
Specifically, for every hidden dimension, we report the minimum/maximum, the $0.01/99.99$ percentiles, the $1/99$ percentiles, and the $25/75$ percentiles.
Figures~\ref{fig:cldc_dist_qk}--\ref{fig:cldc_dist_down} compare the original module inputs with the corresponding representations after applying CLDC for all recurrent loops.

The raw activations exhibit pronounced channel-wise anisotropy and sparse outliers across different module types.
For the attention projections, including $q$, $k$, and $v$, a small subset of channels can reach values several times larger than the central activation range, and these extreme channels become particularly pronounced in later loops.
Similar behavior appears in the MLP gate and up projections.
An especially severe case occurs at the input of the down projection: in the first loop, isolated channels exceed a magnitude of 20 even though the central percentiles remain concentrated close to zero.
Such large discrepancies between a few extreme channels and the majority of activations considerably enlarge the effective quantization range, making uniform low-bit activation quantization inefficient.

After CLDC, the activation distributions become substantially more balanced across hidden dimensions.
The extreme channel-wise spikes are strongly suppressed, while the percentile envelopes become smoother and more homogeneous.
For $q/k/v$ and the MLP gate/up projections, CLDC consistently converts the highly irregular raw distributions into compact ranges across all recurrent loops.
The down projection shows an even more pronounced transformation: the extremely large outliers observed in the raw input are reduced to a compact sub-unit range, with later loops exhibiting particularly narrow distributions.
Although the exact activation range remains module- and loop-dependent, CLDC greatly reduces the disparity between outlier channels and the bulk activation distribution.

These visualizations illustrate that the difficulty of activation quantization in LoopLMs is not only caused by the absolute magnitude of activations, but also by their highly heterogeneous channel-wise geometry and its variation across recurrent computation.
By regularizing these distributions before quantization, CLDC provides a substantially better-conditioned activation space, reducing the influence of isolated outliers on the quantization range while preserving the relative structure of the dominant activation mass.
This observation complements the layer-wise error analysis in Sec.~\ref{app:layerwise_error}: the more balanced intermediate representations produced by CLDC provide a favorable basis for suppressing the accumulation and re-amplification of quantization errors across recurrent loops.

\end{document}